\documentclass[runningheads]{llncs}

\usepackage{eccv}

\usepackage{eccvabbrv}

\usepackage{graphicx}
\usepackage{booktabs}
\usepackage[utf8]{inputenc}
\usepackage{amsmath, amssymb}
\usepackage{mathrsfs}
\usepackage{multirow,multicol}
\usepackage{tabularray}
\usepackage{graphicx}
\usepackage{algorithm}
\usepackage{algorithmic}
\usepackage{wrapfig}
\usepackage[table]{xcolor}
\usepackage{booktabs}      % \toprule \midrule \bottomrule
\usepackage{siunitx}       % S 列、数字对齐
\usepackage{makecell} 
\usepackage{array} 

\usepackage[accsupp]{axessibility}  % Improves PDF readability for those with disabilities.

\usepackage{hyperref}

\usepackage{orcidlink}

\begin{document}

%\title{FlowDec: Temporal Conditional Flow Decorruptor for Robust Vision-Language Navigation} 
\title{FlowDec: Temporal Conditional Flow Decorruptor for Robust Continuous Vision-Language Navigation} 



\author{Yufei Zhang\inst{1}\orcidlink{0009-0009-3863-204X} \and
        Changhao Chen\inst{1}\thanks{Corresponding author.}\orcidlink{0000-0002-8341-6399}}

\authorrunning{Y.~Zhang et al.}

\institute{
  The Hong Kong University of Science and Technology (Guangzhou), China \\
\email{yzhang957@connect.hkust-gz.edu.cn \quad changhaochen@hkust-gz.edu.cn}        
}

\maketitle
\vspace{-1.7em}

\begin{abstract}
Vision-and-Language Navigation in Continuous Environments (VLN-CE) requires agents to follow natural-language instructions in unseen scenes. While Large Models (LMs) have advanced VLN-CE, their performance remains severely degraded by real-world visual corruptions, a critical yet underexplored domain constraint. We introduce Temporal Conditional Flow Decorruptor (FlowDec), a novel image restoration framework tailored for LM-based VLN-CE. FlowDec integrates a hybrid temporal conditioning strategy to align the generative flow path with historical context and employs action-centroid guided filtering to dynamically assess and integrate outputs. Extensive experiments demonstrate that FlowDec outperforms state-of-the-art decorruption methods in both navigation accuracy and generation latency. Our approach establishes a robust, efficient paradigm for resilient embodied navigation in unpredictable real-world conditions. The code is released in \href{https://github.com/FeeFee-1/FlowDec-Temporal-Conditional-Flow-Decorruptor-for-Robust-Continuous-Vision-Language-Navigation}{\textcolor{blue}{this URL}}.
\keywords{Vision Language Navigation \and Conditional Flow Matching \and Image Restoration}
\end{abstract}    
\section{Introduction}
\label{sec:intro}

Vision-and-Language Navigation (VLN) has emerged as a cornerstone problem in Embodied AI, requiring an agent to interpret natural language instructions and execute goal-directed navigation in a 3D environment~\cite{Park_Kim_2023}. Early work focused on discrete simulators with predefined navigation graphs~\cite{anderson2018vision, qi2020reverie, ku2020room, chen2022think}, where action spaces and connectivity were artificially constrained. While such settings enabled rapid algorithmic progress, they abstracted away many of the complexities inherent to real-world deployment. Recent advances have shifted toward VLN in continuous environments (VLN-CE)~\cite{krantz2020beyond}, where agents must generate physically executable trajectories, reason over long temporal horizons, and ground language in raw sensory streams. This paradigm—exemplified by datasets such as R2R-CE and RxR-CE—moves VLN closer to practical robotic systems operating in homes, offices, and outdoor spaces. In this setting, perceptual fidelity and action reliability are not merely performance metrics; they are safety-critical requirements.

%Motivated by the remarkable reasoning ability of Large Models (LMs) — including Large Language Models (LLMs) and Vision-Language Models (VLMs) — recent works have integrated them into VLN-CE~\cite{zhang2024navid, cheng2024navila, zhang2024uni, long2024instructnav}, achieving competitive performance~\cite{kim2024openvla, black2024pi_0, zitkovich2023rt}. However, LM-based VLN systems still face substantial challenges, including the Sim-to-Real gap and limited spatial intelligence~\cite{anderson2021sim, yang2025thinking}. Existing approaches mitigate these limitations by introducing auxiliary tasks to enhance reasoning~\cite{zhang2024navid, zhang2024uni} or scaling real-world data via web and video sources~\cite{cheng2024navila, lin2023learning}. 

The recent integration of Large Models (LMs), including Large Language Models and Vision-Language Models, has significantly advanced VLN-CE~\cite{zhang2024navid, cheng2024navila, zhang2024uni, long2024instructnav,internnav2025,zeng2025janusvln}. By leveraging their strong reasoning and cross-modal alignment capabilities~\cite{kim2024openvla, black2024pi_0, zitkovich2023rt}, LM-based agents such as NaVid \cite{zhang2024navid} demonstrate competitive long-horizon planning and instruction grounding. These approaches suggest a promising route toward general-purpose embodied agents. However, their impressive reasoning abilities often mask a fundamental vulnerability: they implicitly assume clean and stable visual inputs. In practice, this assumption rarely holds.
Real-world navigation inevitably exposes robots to diverse visual corruptions arising from both external and internal factors. Rapid ego-motion induces motion blur; low-light or over-exposure alters luminance statistics; rain, dust, or lens contamination introduce structured noise; and sensor imperfections generate stochastic artifacts. Such corruptions distort critical spatial cues—object boundaries, floor textures, depth discontinuities—that directly affect waypoint prediction and action selection. In safety-critical scenarios, even minor perceptual degradation can cascade into catastrophic failures, including collisions, disorientation, or inability to reach the goal.

\begin{figure}[t]
  \centering
   \includegraphics[width=1\linewidth]{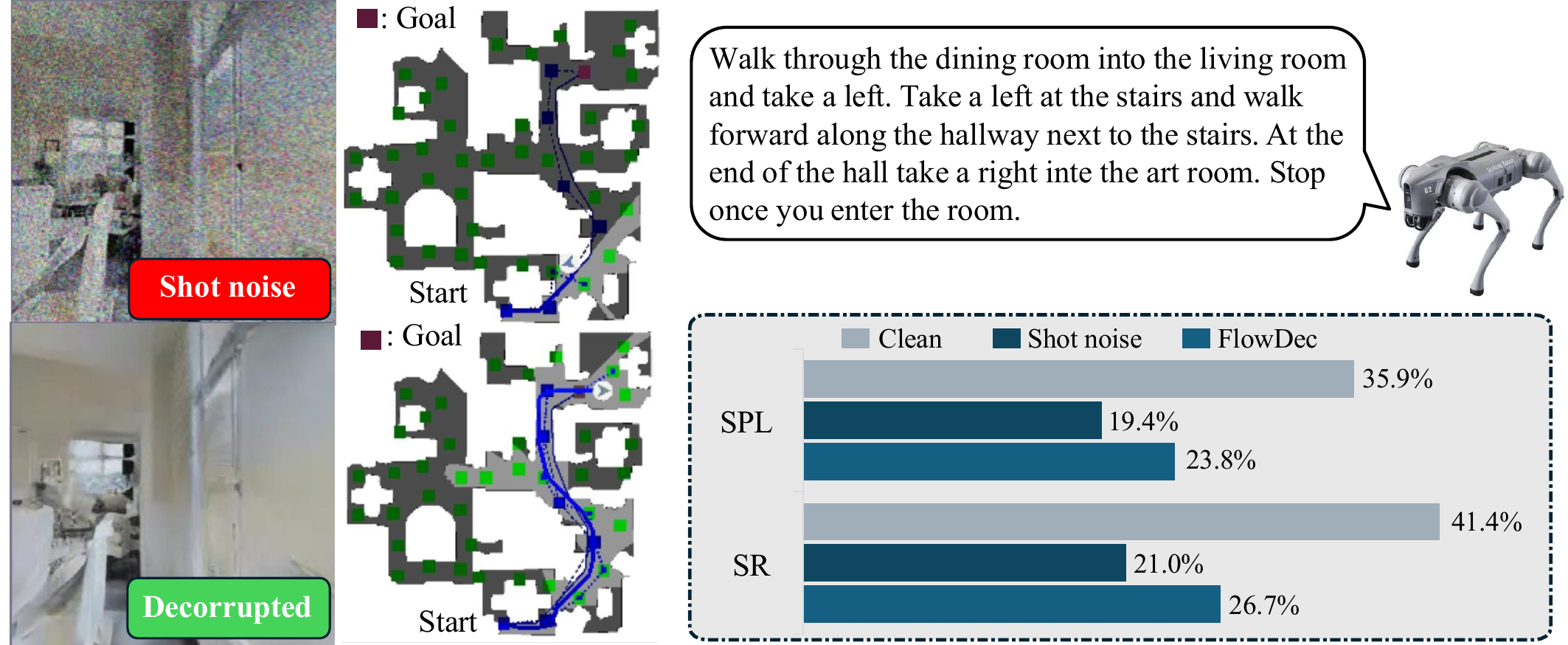}
   \caption{LM-based VLN models suffer from corruption despite their strong generalization ability, while FlowDec can effectively alleviate the influence from corruption. The experiment is conducted on R2R-CE dataset using NaVid~\cite{zhang2024navid} as backbone.}
   \label{fig1}
\end{figure}

 %Despite substantial progress in scaling data and enhancing reasoning, robustness to realistic visual corruptions remains underexplored in VLN-CE.

%To illustrate this vulnerability, we applied shot noise to NaVid \cite{zhang2024navid}, a representative LLM-based VLN model. As shown in Figure~\ref{fig1}, despite extensive training, its success rate drops from 41.4\% to 21.0\%—a 49\% relative drop of 49\%. Corruptions arise from external (weather, luminance) and internal (sensor, motion) factors, far beyond training distributions. This raises a fundamental challenge: How can we achieve robust VLN model under diverse corruptions without corresponding priori knowledge while preserving real-time inference speed?

Despite the growing realism of VLN-CE and the scaling of LM-based agents, robustness to visual corruptions has not been systematically studied in this setting. To our knowledge, this work is the first to explicitly investigate and address the problem of image decorruption for VLN in continuous environments. 
Our empirical analysis reveals the severity of this issue. As shown in Figure~\ref{fig1}, when synthetic shot noise is applied to a representative LM-based VLN agent \cite{zhang2024navid}, its navigation success rate drops by nearly half, despite extensive pretraining. This degradation highlights a broader challenge: how can we endow VLN systems with robustness to unknown and diverse corruptions—without requiring prior knowledge of corruption types, without retraining the navigation backbone, and without sacrificing real-time performance?

% Note that corruptions can be raised from external environmental changes, sensor degradation, or communication failures beyond typical shot noise, highlight the need to enhance the robustness of LM-based VLN models.

%Current solutions fall into two categories to alleviate visual corruptions: blind image denoising (BID) and test-time adaptation (TTA). BID methods recover images from unknown corruptions using subnetworks or adversarial networks \cite{chen2018image, guo2019toward, yue2019variational, zhang2023practical}, while TTA methods usually address this by adjusting model parameters at test time \cite{wang2020tent, niu2022efficient, niu2023towards, wang2022continual}. For LM-based VLN-CE, diffusion-based TTA methods are more effective, as they focus on adjusting input images rather than modifying model parameters, thus avoiding catastrophic forgetting \cite{gao2023back, oh2024efficient, tsai2024gda}. However, these methods primarily process independent images, whereas VLN-CE requires temporal inter-frame consistency and rapid denoising for its online nature, while the downstream models should generate physically meaningful action rather than just denoising image, which may look good but sometimes has no physical meaning.

Addressing this gap is non-trivial. A robust VLN-CE system must handle unknown and diverse corruptions without prior knowledge of their type or intensity, preserve temporally consistent observations across frames, maintain physically meaningful action predictions, and operate under strict real-time constraints. Existing image restoration approaches are not designed for this regime. Blind image denoising methods prioritize perceptual quality rather than downstream navigation consistency \cite{chen2018image, guo2019toward, yue2019variational, zhang2023practical}. Test-time adaptation (TTA) methods update model parameters during inference, introducing the risk of forgetting—especially problematic for large frozen LMs \cite{wang2020tent, niu2022efficient, niu2023towards, wang2022continual}. Diffusion-based TTA approaches mitigate parameter updates by refining inputs, yet they typically process frames independently and emphasize visual realism, overlooking temporal coherence and action-level consistency required for embodied navigation \cite{gao2023back, oh2024efficient, tsai2024gda}. 

To address these limitations, we propose \textbf{FlowDec}, a  Temporal Conditional \textbf{Flow} \textbf{Dec}orruptor framework designed specifically for robust embodied navigation.
 Unlike prior restoration methods developed for image enhancement or video generation, FlowDec is explicitly optimized for navigation-aware, real-time decorruption. Our method introduces two key modules: (1) a hybrid temporal conditioning strategy that dynamically integrates historical frames and condition types to align the generative flow path and enhance temporal consistency, and (2) Action centroids, which model differential feature distributions of atomic actions (\textit{e.g.,} forward, turn left, turn right; predicted by LM-based VLN models \cite{zhang2024navid, cheng2024navila}) to assess output quality and guide adaptive integration of latent reconstructions. This design decouples robustness enhancement from the navigation backbone, making FlowDec broadly compatible with diverse LM-based VLN agents. Moreover, by leveraging flow matching instead of iterative diffusion sampling, FlowDec achieves substantially faster inference, meeting the stringent latency requirements of online robotic deployment.
 %Compared with prior temporal-consistency diffusion methods that target video generation or reconstruction~\cite{jin2024pyramidal,martin2024pnp}, FlowDec emphasizes addressing real-time decorruption for embodied navigation, where outputs must immediately support reliable actions under strict latency constraints.
 We evaluate FlowDec on R2R-CE and RxR-CE under six representative corruption types across more than 5,000 unseen trajectories. FlowDec improves relative navigation success by 25.33\% and 9.38\% on the two benchmarks, respectively, while achieving 3×–8× faster decorruption compared to representative diffusion-based TTA methods. We further validate its effectiveness in real-world robotic experiments, demonstrating consistent gains under practical sensory degradations.
To summarize, our contributions are threefold:
%We evaluate FlowDec on R2R-CE~\cite{krantz2020beyond} and RxR-CE~\cite{ku2020room} under 6 corruption types across 5000+ unseen trajectories. Experiments demonstrate that with the implementation of FlowDec, the model's success rate improved by relatively 25.33\% and 9.38\% across the two datasets. Moreover, FlowDec achieves single-step decorrupting speeds that are 3$\times$ to 8$\times$ speedup over representative diffusion-based TTA methods~\cite{oh2024efficient}. Real-world experiments are also provided to validate FlowDec's effectiveness in real environments. 

\begin{enumerate}
\item We identify corruption robustness as a fundamental yet overlooked bottleneck for VLN in continuous environments and formulate navigation-aware decorruption as a real-time conditional flow-matching problem. %We identify image decorruption as a critical robustness challenge for VLN-CE and introduce FlowDec, a conditional flow-matching decorruptor with hybrid temporal conditioning, enabling consistent and domain-robust restoration for online navigation.  
%We identify corruption robustness as a fundamental yet overlooked bottleneck for LM-based VLN-CE and formulate navigation-aware decorruption as a real-time conditional flow-matching problem.
%We formulate the task of image decorruption in VLN-CE as a critical robustness challenge under diverse and unknown corruptions. To address this, we propose FlowDec, which decorrupts images through a CFM method. FlowDec employs a hybrid training strategy integrating multiple conditioning types in dynamic proportions. Different conditions are designed to share the same optimisation objective, thereby enabling diverse conditions to achieve better flow path alignment during the training phase, thereby enables the model to capture rich historical context while progressively adapting to the domain gap between synthetic outputs and ground-truth images. The resulting decorrupted images exhibit strong spatial consistency within frames and temporal coherence across sequences, significantly enhancing navigation reliability in unseen environments.
\item We propose FlowDec with hybrid temporal conditioning and action-centric guidance, enabling temporally consistent, physically meaningful restoration without modifying the navigation backbone. %We introduce action centroids as guidance to enforce navigation-consistent generation, ensuring that inter-frame changes follow navigation-specific motion patterns while maintaining high efficiency compared with BID and TTA methods.
%FlowDec introduce action centroids which leverage navigation-specific atomic actions to define expected differential feature distributions between consecutive frames. By using these centroids as a quality verification mechanism, FlowDec ensures that inter-frame changes in generated images align with the statistical distribution of corresponding actions. This action-conditioned consistency constraint guarantees temporal fidelity in the restoration process, enabling stable scene interpretation and real-time decision-making in dynamic VLN-CE settings.
\item Extensive experiments demonstrate that FlowDec achieves superior restoration speed and navigation success, establishing a new robustness paradigm for VLN under diverse corruptions.
\end{enumerate}
\section{Related Works}
\label{sec:RW}

\subsection{LM-based VLN Models}
LLMs have advanced embodied AI by enabling strong reasoning and language grounding in robotic control~\cite{zitkovich2023rt, black2024pi_0, kim2024openvla} and discrete VLN planning~\cite{zhou2024navgpt, zhou2024navgpt2, pan2023langnav, chen2024mapgpt}. Recent VLN-CE works integrate LLMs for continuous navigation~\cite{zhang2024navid, zhang2024uni, cheng2024navila, long2024instructnav}. End-to-end models like NaVid series fine-tune VLMs~\cite{chiang2023vicuna} to execute actions from RGB and language inputs~\cite{zhang2024navid,zhang2024uni}, while NaVILA~\cite{cheng2024navila} adopts a hierarchical LLM planner + low-level controller~\cite{miki2022learning} to improve navigation capability. 
LM-based VLN faces (i) limited spatial reasoning ability~\cite{yang2025thinking, zhang2024vision}, and (ii) sim-to-real gaps due to scarce real instruction and sensor data~\cite{savva2019habitat, anderson2021sim, kamath2023new, hong2025general}. Prior works improve grounding via chain-of-thought prompting~\cite{long2024instructnav, zhou2024navgpt, chen2024mapgpt, pan2023langnav} and additional data from video sources~\cite{cheng2024navila, lin2023learning} or auxiliary tasks~\cite{zhang2024navid, zhang2024uni}. Robustness to visual corruptions remains largely unaddressed, despite strong impact in real deployment. Our work directly tackles this gap without retraining the VLN agent.

\subsection{Image Decorruption } %这些方法都不是应用于VLN的我觉得这里写in VLN 不合适？

VLN robustness requires handling unpredictable visual corruptions that cannot be anticipated during training. Two primary paradigms exist: blind image denoising (BID) and test-time adaptation (TTA). BID removes unknown noise by learning to estimate corruption distributions~\cite{chen2018image, guo2019toward, yue2019variational, zhang2023practical}. For example, SCUNet~\cite{zhang2023practical} synthesizes realistic degradations and uses pixel and adversarial losses for robust denoising. However, data-driven BID alone struggles with unseen real-world corruptions. TTA instead adapts to test-time inputs without source data~\cite{liang2025comprehensive, wang2020tent, zhang2025analytic}. Classical TTA methods focus on discrete actions~\cite{gao2023fast, tan2025source, niu2023towards} and are difficult to extend to VLN-CE, where pseudo-labels are ambiguous in continuous spaces and single-sample online adaptation risks catastrophic forgetting~\cite{wang2020tent, gao2023back, wang2022continual}. Diffusion-based TTA offers a promising alternative by correcting input images instead of modifying model parameters~\cite{gao2023back, guo2025everything, oh2024efficient}. DDA~\cite{gao2023back} demonstrates improved robustness across noise types but requires many denoising steps and mainly targets noise rather than diverse corruptions~\cite{tsai2024gda}. Decorruptor~\cite{oh2024efficient} accelerates inference via latent denoising and augmentation but still processes frames independently.
Yet VLN-CE requires real-time generation and temporal consistency across frames to maintain stable navigation. Our method addresses both challenges by improving efficiency and enforcing consistency across sequential observations, providing robust perception for online VLN-CE.

\section{Temporal Conditional Flow Decorruptor}
\label{sec:method}
\begin{figure*}[t]
  \centering
   \includegraphics[width=1\linewidth]{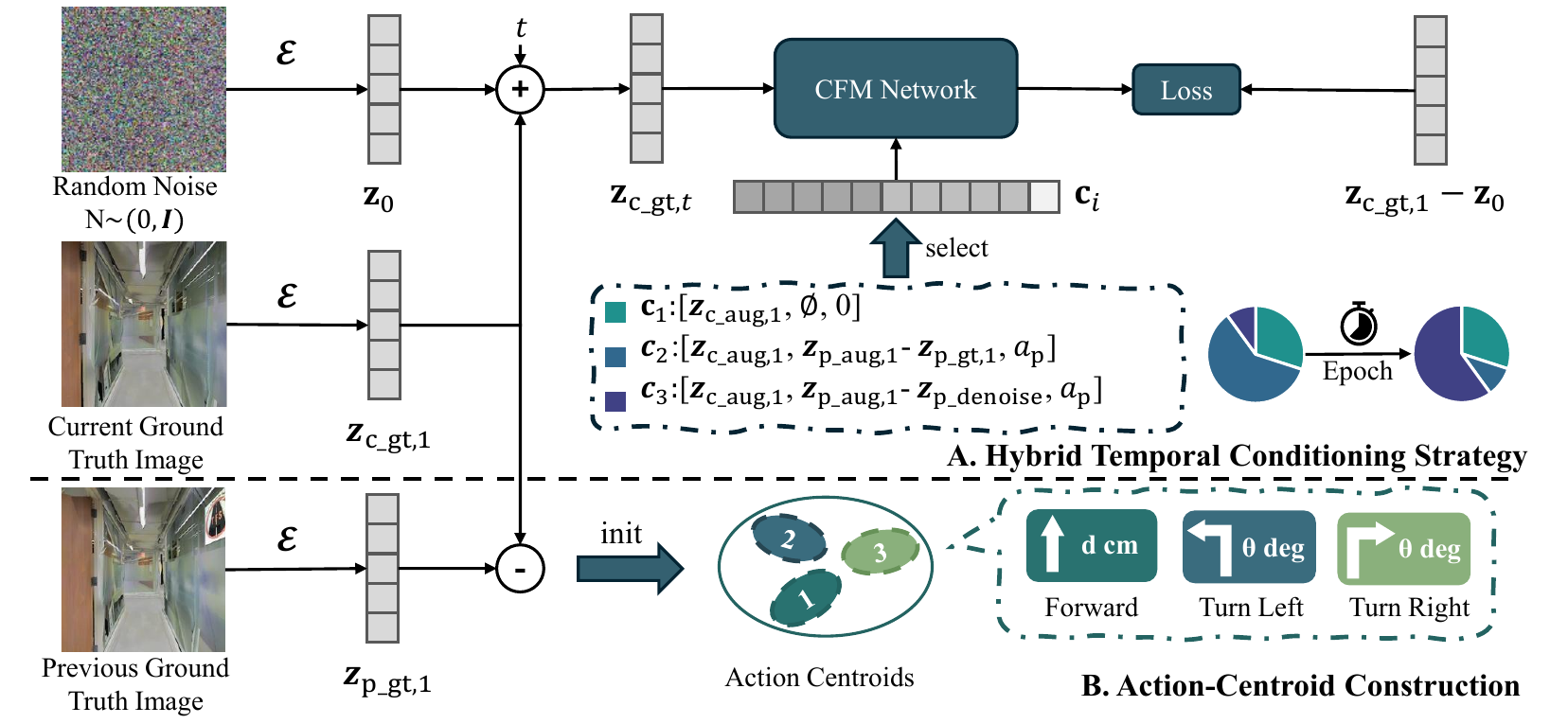}
   \caption{Overview of training phase. \textbf{(A)} Model utilizes three types of conditions for training. The proportion of $\textbf{c}_1$ remains constant, while $\textbf{c}_3$ gradually replaces $\textbf{c}_2$ as the number of epochs increases. \textbf{(B)} Ground-truth image pairs are used to construct action centroids, which capture expected latent differentials per atomic action.}
   \label{fig2}
   % \vspace{-1.0em}
\end{figure*}

This section introduces FlowDec, a novel framework for mitigating image corruption in VLN-CE. FlowDec takes corrupted images and atomic action labels as input and outputs decorrupted images for the VLN agent. It is trained using latent conditional flow matching (CFM) \cite{dao2023flow}, augmented with a hybrid temporal conditioning strategy (Figure~\ref{fig2}) to align the generative flow path with historical information. At inference, action centroids are used to assess inter-frame consistency and dynamically integrate outputs from multiple conditions (Figure~\ref{fig3}). This mechanism ensures that generated images exhibit action-coherent transitions, enhancing spatial and temporal fidelity for robust downstream navigation.

\subsection{Problem Formulation and Preliminaries}
\label{sec:3.1}

The objective of FlowDec is to reconstruct a decorrupted image $\mathbf{x}_{\text{c\_decor}}$ from a corrupted input $\mathbf{x}_{\text{c\_cor}}$ using CFM. Following prior CFM-based generation works \cite{dao2023flow, gode2024flownav}, FlowDec operates in the latent space using variable $\mathbf{z}$, obtained via a pretrained variational autoencoder (VAE) encoder $\mathcal{E}$ and reconstructed with decoder $\mathcal{D}$ \cite{Kingma_Welling_2013}.

 %只有上面这段是Problem Formulation ，感觉没有必要区分subsection
 
Given the ground-truth observation image latent $\mathbf{z}_{\text{c\_gt,1}} = \mathcal{E}(\mathbf{x}_{\text{c\_gt}}) \sim p_1$ and a random noise image latent $\mathbf{z}_0 \sim p_0$, the latent CFM network aims to estimate a coupling $\pi(p_0, p_1)$ that models the evolution between these distributions. This objective can be formulated as solving an ordinary differential equation (ODE):

\begin{equation}
  \text{d}\textbf{z}_{\text{c\_gt},t}=u(\textbf{z}_{\text{c\_gt},t},t)\text{d}t,
  \label{eq1}
\end{equation}
where $t \in [0,1]$ denotes time, $u(\mathbf{z}_{\text{c\_gt},t}, t): [0,1] \times \mathbb{R}^d \to \mathbb{R}^d$ is a time-varying velocity field, and $d$ is the dimension of the latent feature space. By parameterizing the velocity field with a trainable neural network $v_\theta(\mathbf{z}_{\text{c\_gt},t}, t)$, the estimation of $\theta$ reduces to a least-squares regression problem:
\begin{equation}
 \mathcal{L}_{\text{FM}}(\theta) = \ \mathbb{E}_{\textbf{z}_{\text{c\_gt},t},t}\left [ \left \| v_{\theta}\left( \textbf{z}_{\text{c\_gt},t},t\right)-u(\textbf{z}_{\text{c\_gt},t},t) \right \| _{2}^{2}   \right ].   
  \label{eq2}
\end{equation}

To make the velocity field $u(\mathbf{z}_{\text{c\_gt},t}, t)$ tractable, we introduce a conditioning latent variable $\mathbf{l}$ and define a marginal probability path $p_t(\mathbf{z}_{\text{c\_gt},t} \mid \mathbf{l})$ that varies according to $\mathbf{l}$:
\begin{equation}
p_{t}(\textbf{z}_{\text{c\_gt},t}) = \int p_{t}(\textbf{z}_{\text{c\_gt},t} \mid \textbf{l})q(\textbf{l})\text{d}\textbf{l}, 
  \label{eq3}
\end{equation}
where $q(\mathbf{l})$ is a distribution over the conditioning variable. Incorporating this into the flow matching framework, the CFM loss turns to:
\begin{equation}
 \mathcal{L}_{\text{CFM}}(\theta)=\mathbb{E}_{\textbf{z}_{\text{c\_gt},t},\textbf{l},t}\left [ \left \| v_{\theta}\left( \textbf{z}_{\text{c\_gt},t},t\right)-u(\textbf{z}_{\text{c\_gt},t},t \mid \textbf{l}) \right \| _{2}^{2}\right ].
  \label{eq4}
\end{equation}

It has been shown \cite{tong2023improving, lipman2022flow} that optimizing $\mathcal{L}_{\text{FM}}(\theta)$ (Eq.~\ref{eq2}) is equivalent to optimizing $\mathcal{L}_{\text{CFM}}(\theta)$ (Eq.~\ref{eq4}), \textit{i.e.,} $\nabla_\theta \mathcal{L}_{\text{FM}}(\theta) = \nabla_\theta \mathcal{L}_{\text{CFM}}(\theta)$. Thus, for any conditioning variable $\mathbf{l}$, the optimization is valid provided the boundary conditions of $p_t(\mathbf{z}_{\text{c\_gt},t} \mid \mathbf{l})$ are satisfied and the process remains continuous. Following \cite{lipman2022flow, liu2022flow}, we define the conditional probability path as a Gaussian distribution:
\begin{equation}
p_{t}(\textbf{z}_{\text{c\_gt},t} \mid \textbf{l})=\mathcal{N}(\textbf{z}_{\text{c\_gt},t} \mid t\textbf{z}_\text{c\_gt,1}+(1-t)\textbf{z}_0,\sigma^2),
  \label{eq5}
\end{equation}
where, as $\sigma \to 0$, the velocity field becomes constant:
\begin{equation}
u(\textbf{z}_{\text{c\_gt},t},t\mid \textbf{l})=\textbf{z}_\text{c\_gt,1} - \textbf{z}_0,
  \label{eq6}
\end{equation}
and $\mathbf{z}_{\text{c\_gt},t}$ reduces to a linear interpolation between $\mathbf{z}_{\text{c\_gt},1}$ and $\mathbf{z}_0$:
\begin{equation}
\textbf{z}_{\text{c\_gt},t}=t\textbf{z}_\text{c\_gt,1}+(1-t)\textbf{z}_0.
  \label{eq7}
\end{equation}

\subsection{Temporal Conditional Flow Matching Learning Strategy} %我还是觉得strategy比较适合
\label{sec:3.2}
Robust viewpoint recovery in sequential navigation demands temporal continuity and resilience to real-world corruptions. Standard flow matching on single images ignores inter-frame relationships and struggles with unseen degradations. While naive augmentation risks overfitting and fails to promote spatial understanding. Therefore simply apply these approaches are suitable for VLN-CE tasks. We propose a hybrid temporal conditioning strategy to align the generative flow path with historical context.

\textbf{CFM Baseline.}
Building on the flow matching framework outlined in Section~\ref{sec:3.1}, the loss for CFM with a constant velocity field is expressed as:
\begin{equation}
 \mathcal{L}_{\text{CFM}}(\theta)=\mathbb{E}_{\textbf{z}_{\text{c\_gt},t},t}\left [ \left \| \textbf{z}_\text{c\_gt,1}-\textbf{z}_0-v_{\theta}\left( \textbf{z}_{\text{c\_gt},t},t\right)\right \| _{2}^{2}\right ],
  \label{eq8}
\end{equation}
where $  \mathbf{z}_{\text{c\_gt},1} = \mathcal{E}(\mathbf{x}_{\text{c\_gt}})  $ is the latent representation of the ground-truth image, $\mathbf{z}_0$ is the random noise latent, and $  v_\theta  $ is the parameterized velocity field.

During inference, the corrupted image latent $\mathbf{z}_{\text{c\_cor},1} = \mathcal{E}(\mathbf{x}_{\text{c\_cor}})$ serves as conditional information $\mathbf{c}_i$, where $i$ denotes different condition types. The CFM loss is thus reformulated as:
\begin{equation}
 \mathcal{L}_{\text{CFM}}(\theta)=\mathbb{E}_{\textbf{z}_{\text{c\_gt},t},t}\left [ \left \| \textbf{z}_\text{c\_gt,1}-\textbf{z}_0-v_{\theta}\left( \textbf{z}_{\text{c\_gt},t},\textbf{c}_i,t\right)\right \| _{2}^{2}\right ].
  \label{eq9}
\end{equation}

\textbf{Hybrid Temporal Conditions.}
Since corrupted images are inaccessible during training, we adopt the corruption modeling scheme from \cite{oh2024efficient} to enhance the robustness of the FlowDec against unseen corruptions. This approach uses augmented images to simulate corruption, training the model to restore these images to their clean counterparts. We employ a hybrid augmentation strategy combining PIXMIX \cite{Hendrycks_2022} and SimSiam \cite{Chen_He_2021}, consistent with \cite{oh2024efficient}. We introduce a hybrid temporal conditioning strategy through $\mathbf{c}_i$, which is defined as:
\begin{equation}
\textbf{c}_i = \begin{cases}
[ \textbf{z}_{\text{c\_aug},1}, \emptyset,0] &,i = 1 \\
[\textbf{z}_{\text{c\_aug},1}, \textbf{z}_{\text{p\_aug},1} - \textbf{z}_{\text{p\_gt},1},a_p] &,i = 2 \\
[\textbf{z}_{\text{c\_aug},1}, \textbf{z}_{\text{p\_aug},1} - \textbf{z}_{\text{p\_denoise}},a_p] &,i = 3
\end{cases},
\label{eq10}
\end{equation}
where $\mathbf{z}_{\text{c\_aug},1}$ is the latent of the current augmented image, $\mathbf{z}_{\text{p\_aug},1}$ and $\mathbf{z}_{\text{p\_gt},1}$ are the latents of the previous augmented and ground-truth images, respectively, $\mathbf{z}_{\text{p\_denoise}}$ is the denoised latent from the previous frame, and $a_p$ denotes the atomic action taken by the agent, as provided by LM-based navigation models \cite{zhang2024navid, cheng2024navila}. The action $a_p$ represents specific action with fixed movement amplitudes, capturing differential changes between consecutive image frames.

\textbf{Condition Design and Training Schedule.}
Conditions in $\mathbf{c}_i$ correspond to different intensions. For $\mathbf{c}_1$, solely using current augmented image’s latent features, represents the most commonly employed denoising method. While $\mathbf{c}_2$ and $\mathbf{c}_3$ incorporate information from the previous frame, enabling the model to leverage temporal context. Specifically, $\mathbf{c}_2$ approximates corruption features by computing the difference between the augmented and ground-truth latent of the previous frame, guiding the model toward effective image restoration. Since ground-truth images are unavailable during inference and the synthetic domain of generative models differs from that of real images~\cite{tsai2024gda, guo2025everything}, $\mathbf{c}_3$ uses the denoised latent from the previous frame to progressively bridge the domain gap between synthetic and real images. 

During training, the three conditions are combined in a dynamic proportion. The weight of $\mathbf{c}_1$ remains constant, while $\mathbf{c}_2$ dominates in the early epochs to establish robust corruption modeling. As training stabilizes, the proportion of $\mathbf{c}_2$ is gradually reduced in favor of $\mathbf{c}_3$, allowing the model to adapt to its own domain gap.

\subsection{Action-Centroid Guided Latent Filtering}
\label{sec:3.3}
\begin{figure*}[t]
  \centering
   \includegraphics[width=1\linewidth]{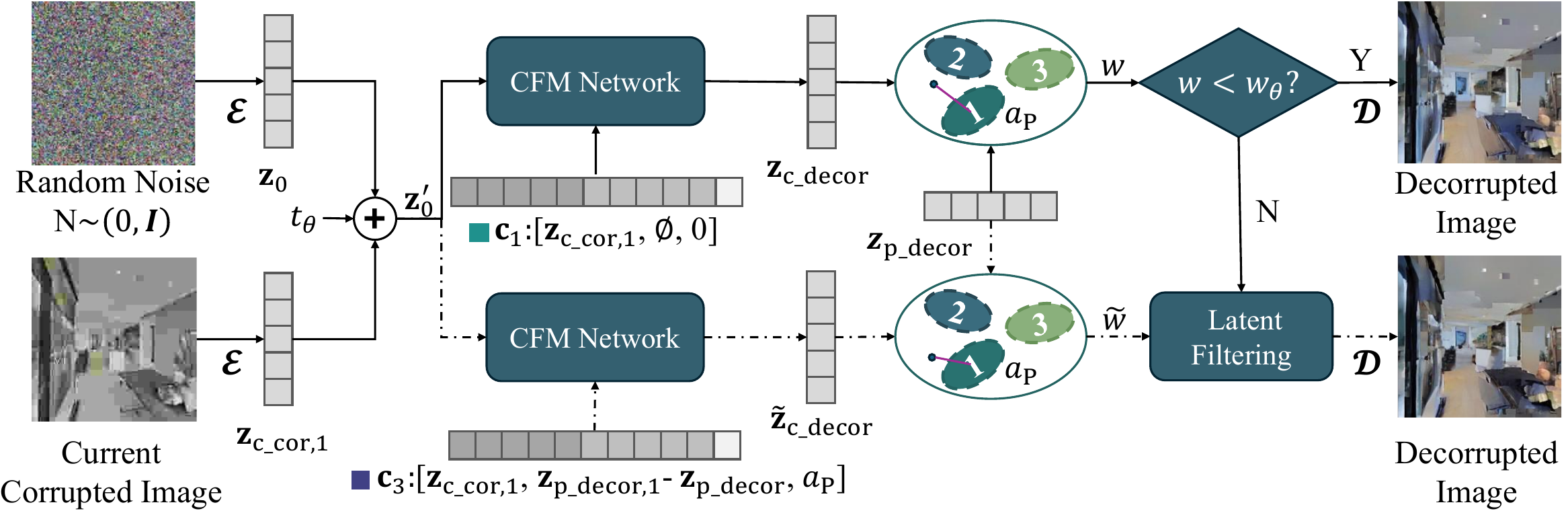}
   \caption{Overview of the inference phase. The model primarily generates using condition $\mathbf{c}_1$. The auxiliary condition $\mathbf{c}_3$ is invoked only when the distance to the corresponding action centroid exceeds a threshold $w_\theta$.}
   \label{fig3}
\end{figure*}

Hybrid temporal conditioning enables dual inference paths using $  \mathbf{c}_1  $ (single-frame) and $  \mathbf{c}_3  $ (temporal). However, always using $  \mathbf{c}_1  $ discards valuable prior-frame context, while over-relying on $  \mathbf{c}_3  $ risks error accumulation during iterative inference—a critical failure mode in denoising.
Our key insight: inter-frame consistency is tightly coupled with atomic actions. We thus propose action-centroid guided corruption filtering, which dynamically evaluates and fuses outputs from $  \mathbf{c}_1  $ and $  \mathbf{c}_3  $ based on action-specific latent distributions. This mechanism selectively leverages temporal cues while preserving single-frame stability, ensuring robust and consistent decorruption in sequential navigation (see Figure~\ref{fig3}).

\textbf{Action-Centroid Construction.}
To exploit navigation dynamics, we treat atomic actions as consistency priors. During training, we compute differential latents $\mathbf{z}_{\text{p\_gt},1} - \mathbf{z}_{\text{c\_gt},1}$ between consecutive ground-truth frames and record per-action mean and variance (Figure~\ref{fig2}). This defines the action centroid:
\begin{equation}
A_{\text{gt},a}=\mathcal{N}(\mu_{\text{gt},a},\sigma_{\text{gt},a}),
\label{eq11}
\end{equation}
where $a$ denotes the type of atomic action. This centroid captures the expected latent differences associated with specific actions, facilitating consistent image reconstruction in continuous navigation tasks.

In sequential image generation, temporal conditioning improves model optimization and domain adaptation but can accumulate errors during iterative inference. To maintain stable and reliable outputs, we need a mechanism to selectively leverage temporal cues without propagating errors. We propose action-guided corruption filtering to balance temporal guidance and single-frame stability for robust latent reconstruction.

\textbf{Model Inference and Latent Filtering.}
Although temporal conditions ($\mathbf{c}_2$, $\mathbf{c}_3$) provide valuable historical context, they risk accumulating generation errors during inference—particularly in iterative denoising during inference period, which is more lethal for iterative generative models. Therefore, we designate $\mathbf{c}_1$ as the primary condition and $\mathbf{c}_3$ as the auxiliary condition. Note that Reducing temporal conditioning reliance at inference does not diminish its training value. Since all conditioning objectives are aligned, flow path trained using temporal conditions also benefits $\mathbf{c}_1$ generation consistency (see validation in Section~\ref{sec:4.3}). 

The inference procedure is illustrated in Figure~\ref{fig3}. The model primarily generates decorrupted latent $\mathbf{z}_{\text{c\_decor}}$ using condition $\mathbf{c}_1$. For the first frame, $\mathbf{z}_{\text{c\_decor}}$ is output directly. For subsequent frames, the quality of the generated latent is verified using action centroids. Specifically, we compute the mean Mahalanobis distance $w$ between the differential features of the current and previous decorrupted latent ($\mathbf{z}_{\text{p\_decor}} - \mathbf{z}_{\text{c\_decor}}$) and the corresponding action centroid distribution, assuming all the dimensions are independent:
\begin{equation}
w=\sqrt{ {\textstyle \sum_{k=1}^{d}\frac{\left ( \textbf{z}_{\text{p\_decor},k}-\textbf{z}_{\text{c\_decor},k}- \mu_{\text{gt},a_p,k} \right )^2 }{d\sigma_{\text{gt},a_p,k}^2 } } } ,
\label{eq12}
\end{equation}
where $d$ is the size of latent dimension, $\mu_{\text{gt},a_p,k}$ and $\sigma_{\text{gt},a_p,k}^2$ are the mean and variance of the action centroid for action $a_p$. If $w$ is below a predefined threshold $w_\theta$, the output is $\mathbf{z}_{\text{c\_decor}}$ from $\mathbf{c}_1$. Otherwise, model needs to decorrupt for the second round and calculate weight $\tilde{w}$ using $\mathbf{c}_3$. If $\tilde{w}<w$, the output result is represented as a weighted combination of $\mathbf{z}_{\text{c\_decor}}$ and $\tilde{\mathbf{z}}_{\text{c\_decor}}$:
\begin{equation}
\mathcal{E}(\textbf{x}_\text{c\_decor}) = \begin{cases}
\frac{\tilde{w}}{\tilde{w}+w} \textbf{z}_{\text{c\_decor}}+\frac{w}{\tilde{w}+w} \tilde{\textbf{z}}_{\text{c\_decor}} &,\tilde{w}<w \\
\textbf{z}_{\text{c\_decor}} &,otherwise
\end{cases}.
\label{eq13}
\end{equation}

This design leverages the inherent stability of images generated with $\mathbf{c}_1$, invoking $\mathbf{c}_3$ only when current frame significantly deviates from previous frame, therefore balances inter-frame consistency with computational efficiency.

Typically, generation begins from random noise $\mathbf{z}_0$ and follows a complete integration path to $\mathbf{z}_1$. While we modify the starting point to further enhance inference efficiency: 
\begin{equation}
{\textbf{z}}'_0=t_\theta\textbf{z}_0+(1-t_\theta)\textbf{z}_{\text{c\_cor},1}.
\label{eq14}
\end{equation}

Experiment in section~\ref{sec:4.3} shows that This modification not only accelerates generation but also reduces randomness in the output, improving performance in different corruption scenarios.
\section{Experiments}
\label{sec/4_Experiment}
In this section, we mainly investigate three questions: \textbf{Q1}: How do different corruptions affect LM-based VLN models?
\textbf{Q2}: Why does FlowDec outperform other SOTA decorruption models? \textbf{Q3}: Are all components of FlowDec necessary for its performance?
\subsection{Implementation Details, Experiment Settings and Baselines}

\textbf{Implementation Details.} 
In FlowDec, the conditional flow matching (CFM) module adopts a U-Net architecture following~\cite{dao2023flow}. The VAE encoder and decoder are implemented based on the original designs in~\cite{dao2023flow,yue2019variational}. We train FlowDec for 20 epochs with a learning rate of $1\times10^{-5}$. The weighting coefficient $w_\theta$ is set to 0.25, and the time threshold $t_\theta$ is set to 0.95.
FlowDec is a model-agnostic visual module that operates independently of the underlying navigation architecture. For evaluation, we adopt NaVid \cite{zhang2024navid} as the baseline, as it is widely adopted and representative among LM-based VLN frameworks~\cite{cheng2024navila,zhang2024uni,zeng2025janusvln,internnav2025}. All training and experiments are conducted on a single NVIDIA RTX 4090 GPU.

To ensure robustness under high intra-action variance across diverse geometric environments, we aggregate statistics from 2,440,814 image pairs collected from over 70 distinct scenes derived from the training splits of R2R-CE and RxR-CE. This large-scale aggregation enables the estimation of empirically stable action centroids with reliable covariance statistics. Additional implementation details are provided in the supplementary material.

\textbf{Experiment Settings.} Simulations are conducted using the Habitat platform \cite{savva2019habitat} on the R2R-CE \cite{krantz2020beyond} and RxR-CE \cite{ku2020room} datasets, comprising 1,839 and 3,669 episodes respectively, across over 50 scenes. Following \cite{wang2022continual, oh2024efficient}, we construct 12 corruption types to assess the robustness of the backbone. Performance is measured using Success Rate (SR) and Success weighted by Path Length (SPL), while SPL is the primary metric due to its balance of accuracy and efficiency \cite{anderson2018evaluation}.

Real-world experiments are performed on the Unitree GO2 robot under four scenes, with 20 trials per setting (see Figure~\ref{fig5}). An episode is successful if the agent stops within 3 m (simulation) or 0.5 m (real-world) of the goal. No real-world data is used for further training or fine-tuning.

\textbf{Baselines.} We compare FlowDec against following methods:
\begin{itemize}
    \item SCUNet \cite{zhang2023practical}: a representative BID approach.
    \item Dec-DPM and Dec-CM \cite{oh2024efficient}: SOTA diffusion-based TTA using diffusion probabilistic models (5 NFEs) and consistency models (2 NFEs).
\end{itemize}
For fair comparison, Dec-DPM and Dec-CM are trained using the same data and augmentation strategies as FlowDec (PIXMIX \cite{Hendrycks_2022} + SimSiam \cite{Chen_He_2021}). NFE settings are chosen to balance performance and inference speed.
\begin{table*}[t]
\centering
\caption{Performance of NaVid under different corruption types on R2R-CE and RxR-CE datasets. All metrics are reported in \%.}
\label{tab:corruption}
\resizebox{1\linewidth}{!}{
\begin{tabular}{cc*{14}{c}}
\toprule
Dataset & Metric & Clean & Gauss. & Shot & Impul. & Defoc. & Motion. & Snow & Fog & Lightout & Pixel. & JPEG. & Cont. & Bright. \\
\midrule
\multirow{2}{*}{R2R-CE} 
 & SR  & 41.40 & 21.42 & 20.99 & 24.63 & 33.06 & 26.64 & 21.26 & 17.29 & 29.80 & 29.82 & 27.95 & 20.55 & 34.48 \\
 & SPL & 35.85 & 18.93 & 19.42 & 22.56 & 29.83 & 23.70 & 18.43 & 14.33 & 26.65 & 25.93 & 25.10 & 18.43 & 31.65 \\
\midrule
\multirow{2}{*}{RxR-CE} 
 & SR  & 45.65 & 31.33 & 30.12 & 34.64 & 40.56 & 42.27 & 30.40 & 29.15 & 39.36 & 39.14 & 34.24 & 29.25 & 40.53  \\
 & SPL & 37.23 & 25.70 & 23.98 & 27.77 & 33.51 & 34.10 & 24.21 & 22.93 & 32.25 & 31.82 & 29.53 & 23.49 & 33.00 \\
\bottomrule
\end{tabular}}\vspace{-0.5em}
\end{table*}
\begin{table*}[t]
\centering
\caption{Performance of different methods under corruption on R2R-CE and RxR-CE. All metrics are reported in \%. The best results are highlighted in \textbf{bold}.}
\label{tab:compare}
\sisetup{
  round-mode = places,
  round-precision = 2,
  zero-decimal-to-integer = false,
  group-digits = false,
}
\newcommand{\best}[1]{%
  \ifdim #1pt=\maxofcol pt \textbf{#1}\else #1\fi
}
\resizebox{\linewidth}{!}{%
\begin{tabular}{
  l c
  *{7}{S[table-format=2.2]}
  >{}S[table-format=2.2]
  c@{\hspace{1em}}
  *{7}{S[table-format=2.2]}
  >{}S[table-format=2.2]
}
\toprule
\multirow{2}{*}{Method} & \multirow{2}{*}{Metric}
 & \multicolumn{7}{c}{R2R-CE} && \multicolumn{7}{c}{RxR-CE} \\
\cmidrule(lr){3-9} \cmidrule(lr){10-17}
 & & {Gauss.} & {Shot}  & {Snow} & {Fog} & {JPEG.} & {Cont.}& {Avg.}
 & & {Gauss.} & {Shot} & {Snow} & {Fog} & {JPEG.} & {Cont.}& {Avg.} \\ \midrule
\multirow{2}{*}{NaVid}
 & SR &21.42&20.99&21.26&17.29&27.95&20.55&21.58&&31.33&30.12&30.40&29.15&34.24&29.25&30.75 \\
 & SPL&18.93&19.42&18.43&14.33&25.10&18.43&19.11&&25.70&23.42&24.21&22.93&\textbf{29.53}&23.49&24.88 \\
\addlinespace
\multirow{2}{*}{Dec-DPM}
 & SR &22.62&22.84&20.71&24.90&22.73&29.85&23.94&&24.48&25.43&19.27&25.94&24.97&29.95&25.01 \\
 & SPL&18.53&18.64&16.13&20.16&18.43&24.84&19.46&&17.46&18.28&13.19&19.09&17.86&22.70&18.10 \\
\addlinespace
\multirow{2}{*}{Dec-CM}
 & SR &\textbf{28.53}&26.35&\textbf{23.27}&15.71&22.19&29.69&24.29&&29.74&26.82&\textbf{31.34}&22.68&31.51&32.60&29.12 \\
& SPL &\textbf{24.22}&22.62&\textbf{20.69}&13.46&20.86&27.24&21.52&&23.13&20.88&\textbf{24.30}&16.98&25.85&26.23&22.90 \\
\addlinespace
\multirow{2}{*}{SCUNet}
& SR & 25.77 & 24.25& 6.87 & 12.32 & 18.81 & 0.05 &14.68&&31.21&30.25&17.88&20.69&30.88&6.73&22.94\\
& SPL& 23.26 & 21.50& 6.47 & 12.11 & 17.83 & 0.04 &13.54&&24.18&23.38&14.82&15.93&26.22&6.66&18.53 \\
\addlinespace
% \multirow{3}{*}{FlowDec}
\multirow{2}{*}{FlowDec} 
& SR  &26.54&\textbf{26.70}&21.40&\textbf{26.87}&\textbf{29.85}&\textbf{30.90}&\textbf{27.04}&&\textbf{32.33}&\textbf{31.35}&30.54&\textbf{35.93}&\textbf{35.76}&\textbf{35.88}&\textbf{33.63} \\
& SPL &23.53&\textbf{23.76}&18.38&\textbf{22.26}&\textbf{26.48}&\textbf{27.32}&\textbf{23.62}&&\textbf{26.41}&\textbf{23.81}&24.17&\textbf{26.14}&29.20&\textbf{28.14}&\textbf{26.31}\\
\bottomrule
\end{tabular}%
}\vspace{-0.5em}
\end{table*}

%\subsection{Performance Comparison}
\subsection{Robustness Analysis of VLN Models} 
\label{sec:4.2}
Table~\ref{tab:corruption} reveals significant performance degradation under various corruptions. Among them, luminance variance (brightness and light-out) and image blur (motion blur and defocus blur) have minimal impact, while fog and snow cause severe drops. This disparity stems from domain mismatch in the training data: lighting variations and motion blur are well-represented in the source domain, enabling effective generalization. In contrast, weather effects (\textit{e.g.,} Fog, Snow) and structured noise (\textit{e.g.,} Gaussian) lie outside the training domain, despite their real-world prevalence, resulting in performance decline. Collecting real-world data spanning such out-of-domain corruptions at scale is infeasible, highlighting the critical need for dedicated decorruption modules.
%Similar trends are also observed on NaVILA~\cite{cheng2024navila} (see supplementary material).

\begin{figure*}[t]
  \centering
  \includegraphics[width=1\linewidth]{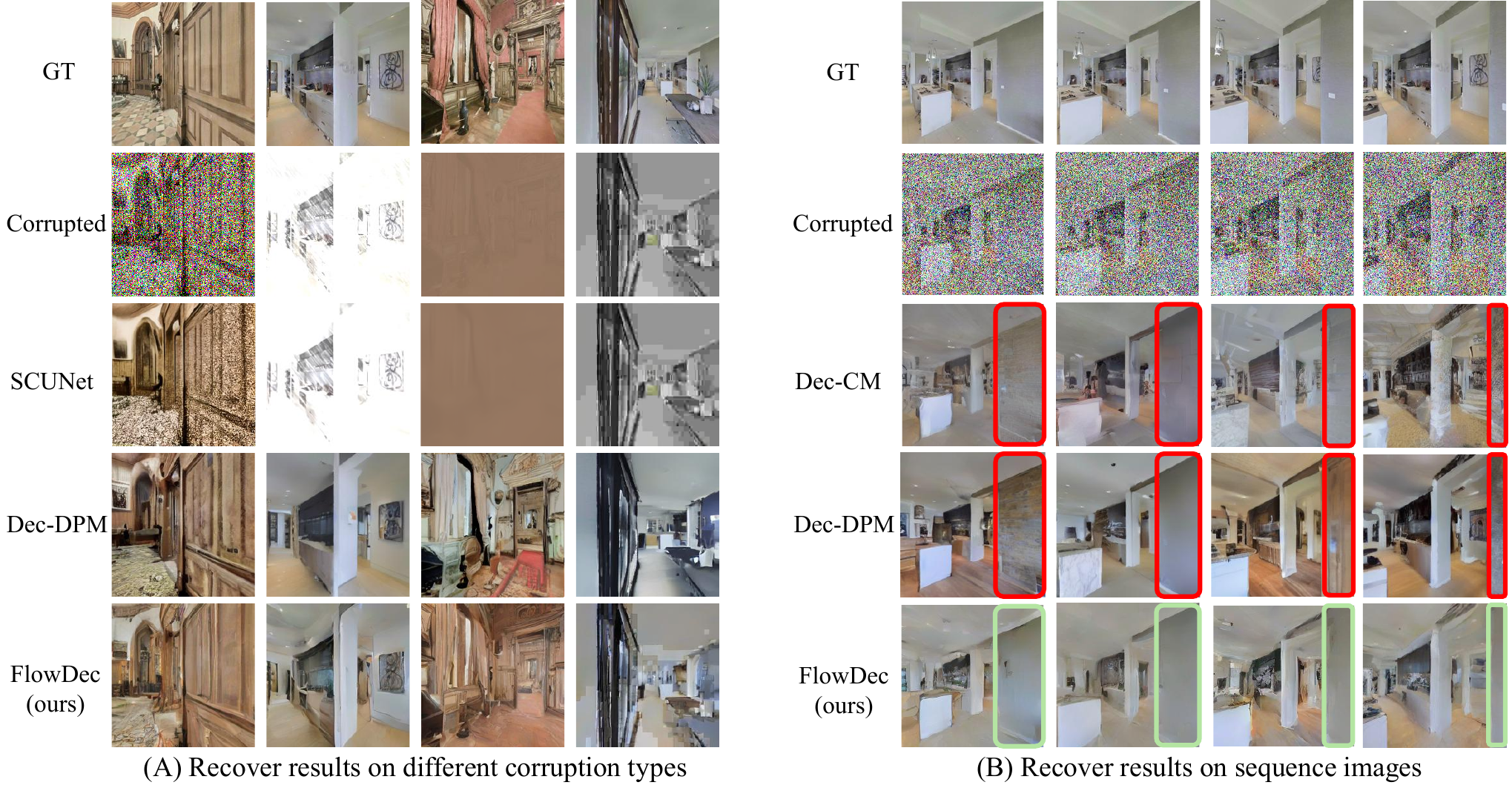}
  % \caption{Illustration of recovered images. (A) Recovery performance of different models under Gaussian noise, Snow, Contrast, and JPEG compression. (B) Recovery performance of different models for the same trajectory under Shot noise. Note that for the wall marked within the frame, only our metho yields results with high  consistency.}
    \caption{Illustration of recovered images. (A) Recovery performance of different models under Gaussian noise, Snow, Contrast, and JPEG compression. (B) Recovery performance of different models for the same trajectory under Shot noise. Note in particular that for the same wall (marked within the frame), only our method yields results with high visual consistency.}
  \label{fig4}
\end{figure*}

\subsection{Performance Evaluation in R2R-CE and RxR-CE Navigation} 
\label{sec:4.3}
In this section, we primarily focus on results for the 6 most destructive corruptions in Table~\ref{tab:corruption}, as these corruptions clearly reveal failure modes and limitations of baseline methods. Minor-impact corruptions are omitted for clarity. 

% To enable fair and representative benchmarking, we evaluate FlowDec against baseline approaches across these corruption categories on both R2R-CE and RxR-CE datasets. This involves a total of 5,508 evaluation episodes spanning more than 50 distinct scenes. For each atomic action, the corrupted image is reconstructed via FlowDec and then passed to the NaVid agent. Final performance metrics are reported as dataset-wide averages.

As evidenced in Table~\ref{tab:compare}, FlowDec consistently surpasses all baselines in aggregate performance, yielding relative SR gains of 25.33\% and 9.37\% over the vanilla model on R2R-CE and RxR-CE, respectively. The BID-based method demonstrates utility solely under Gaussian and shot noise; while in other corruptions its performance either degrades or collapses entirely. This situation is also confirmed in Figure~\ref{fig4}(A) where denoising outcomes are illustrated across diverse corruptions. TTA approaches like Dec-DPM and Dec-CM on the other side, exhibit satisfactory denoising capabilities while their performance is generally inferior to FlowDec, and this gap becomes wider in RxR-CE.

% Although diffusion-based TTA approaches exhibit competitive denoising in isolated scenarios, their efficacy diminishes markedly in others. Moreover, Dec-DPM and Dec-CM exhibit a stronger bias toward exploration than instruction-following. Such bias becomes more obvious In RxR-CE, where actionable space is larger, leading to a sharper drop in SR and SPL. This inconsistency poses a critical limitation for reconstruction-centric frameworks. 
% Similarly, the BID-based method demonstrates utility solely under Gaussian and shot noise; in the presence of alternative corruptions, its performance either degrades or collapses entirely. Figure~\ref{fig4}A provides qualitative insights, which illustrates denoising outcomes across diverse corruptions. Notably, SCUNet's performance align closely with perceptual quality, reinforcing the reliability of our evaluation protocol.

\begin{table*}[t]
\centering
\caption{Performance of different methods using warp error($\downarrow$) on R2R-CE and RxR-CE. The best results are highlighted in \textbf{bold}.}
\label{tab:we}
\resizebox{1\linewidth}{!}{%
\renewcommand{\arraystretch}{0.92}
\begin{tabular}{lccccccccccccccccc}
\toprule
\multirow{2}{*}{Method} & \multicolumn{7}{c}{R2R-CE} && \multicolumn{7}{c}{RxR-CE} \\
\cmidrule(lr){2-8} \cmidrule(lr){9-16}
& {Gauss.} & {Shot}  & {Snow} & {Fog} & {JPEG.} & {Cont.}& {Avg.}& & {Gauss.} & {Shot} & {Snow} & {Fog} & {JPEG.} & {Cont.}& {Avg.} \\ \midrule
NaVid & 0.47&0.52&0.12&0.28&\textbf{0.09}&\textbf{0.03}&0.25& &0.47&0.51&0.14&0.28&\textbf{0.11}&\textbf{0.04}&0.26 \\
\addlinespace
Dec-DPM & 0.30&0.29&0.34&0.33&0.36&0.27&0.31& &0.32&0.31&0.35&0.34&0.37&0.29&0.33 \\
\addlinespace
Dec-CM &0.15&0.15&0.19&0.34&0.13&0.15&0.18& &0.16&0.17&0.22&0.34&0.16&0.16&0.20  \\
\addlinespace
SCUNet& 0.12&0.15&\textbf{0.11}&0.27&\textbf{0.09}&\textbf{0.03}&0.13& &0.14&0.17&\textbf{0.13}&0.28&\textbf{0.11}&\textbf{0.04}&0.14\\
\addlinespace
\rowcolor{gray!45} FlowDec & \textbf{0.10}&\textbf{0.11}&0.12&\textbf{0.15}&0.10&0.10&\textbf{0.11}& &\textbf{0.12}&\textbf{0.13}&0.14&\textbf{0.18}&0.12&0.12&\textbf{0.13}\\
\bottomrule
\end{tabular}%
}\vspace{-0.5em}
\end{table*}
\begin{table*}[t]
\centering
\begin{minipage}{0.21\textwidth}
\centering
\caption{Comparison of model inference time.}
\label{tab:time}
\resizebox{\linewidth}{!}{%
\begin{tabular}{lc}
\toprule
Method & Time (ms) \\
\midrule
NaVid & 370 \\
Dec-DPM & 370+1411 \\
Dec-CM & 370+588 \\
\rowcolor{gray!45} FlowDec & 370+173 \\
\bottomrule
\end{tabular}%
}
\end{minipage}%
\hfill
\begin{minipage}{0.35\textwidth}
\centering
\caption{Comparison on pristine conditions. All metrics are reported in \%.}
\label{tab:clean}
\resizebox{\linewidth}{!}{%
\begin{tabular}{lccc}
\toprule
Method & Metrics & R2R-CE & RxR-CE \\
\midrule
NaVid & SR  & 41.40 & 45.65 \\
\rowcolor{gray!45} FlowDec & SR  & 40.89 & 45.41 \\
NaVid & SPL & 35.85 & 37.23 \\
\rowcolor{gray!45} FlowDec & SPL & 35.96 & 37.03 \\
\bottomrule
\end{tabular}%
}
\end{minipage}
\hfill
\begin{minipage}{0.39\textwidth}
\centering
\caption{Performance of model on real-world experiments. Each task is tested for 20 times.}
\label{tab:real}
\resizebox{\linewidth}{!}{
\begin{tabular}{lcccc}
\toprule
\multirow{2}{*}{Method} & \multicolumn{2}{c}{Indoor} & \multicolumn{2}{c}{Outdoor}\\
\cmidrule(lr){2-3} \cmidrule(lr){4-5}
& Task1 & Task2& Task3 & Task4 \\
\midrule
NaVid   &  0.20  & 0.05 & 0.10 & 0.35 \\
\rowcolor{gray!45} FlowDec&  0.35  & 0.20 & 0.30 & 0.40\\
\bottomrule
\end{tabular}
}
\end{minipage}
\vspace{-1em}
\end{table*}

Considering that Dec-DPM and Dec-CM are distilled from Instruct-Pix2Pix~\cite{brooks2023instructpix2pix}, a diffusion model trained on 45 million image-text pairs, while FlowDec solely utilizes R2R-CE and RxR-CE training splits for training. We attribute FlowDec's superior performance to inter-frame consistency.
% These models inherently possess strong semantic priors and are therefore expected to excel at image restoration. 
% However, FlowDec outperforms both on average using solely the R2R-CE and RxR-CE training splits, attributable to its enforced intra-frame coherence. 
Figure~\ref{fig4}(B) further elucidates this advantage by denoising consecutive frames along identical trajectories. While Dec-DPM and Dec-CM produce visually plausible per-frame outputs, they exhibit pronounced temporal inconsistencies (\textit{e.g.,} walls highlighted in red frames). In contrast, FlowDec leverages hybrid temporal conditioning and action centroid guided filtering to ensure coherent generation, thereby facilitating more reliable scene interpretation by the downstream LM and enhancing navigation fidelity. Given that contemporary LM-driven VLN agents infer ego-localization and command compliance from image histories, inter-frame consistency emerges as a more decisive factor than isolated reconstruction quality. 

%\textcolor{blue}{To further demonstrate FlowDec's ability to generate temporally consistent images, we randomly sample 100 test trajectories from the R2R-CE and RxR-CE respectively, and evaluate using warp error. Unlike semantic-focused metrics (\textit{e.g.,} LPIPS), warp error specifically quantifies motion consistency between consecutive frames and remains robust to corruption-induced semantic distortions. As reported in Table~\ref{tab:we}, FlowDec achieves the lowest warp error on average. It shows particularly large gains under motion-disrupting corruptions such as Gaussian and Shot noise. For corruptions with minimal impact on apparent motion, semantic restoration gains greater importance (\textit{e.g.,} Contrast). However, when warp error increases substantially, navigation performance degrades markedly — even in cases where semantic recovery remains satisfactory (\textit{e.g.,} Dec-DPM under JPEG compression and Dec-CM under Fog condition).}
To further evaluate FlowDec’s temporal consistency, we randomly sample 100 test trajectories from R2R-CE and RxR-CE and measure performance using warp error. Unlike semantic-oriented metrics (\textit{e.g.,} LPIPS), warp error directly quantifies motion consistency between consecutive frames and is less sensitive to corruption-induced semantic distortions.

As shown in Table~\ref{tab:we}, FlowDec achieves the lowest average warp error across datasets and corruption types, with particularly significant improvements under motion-disruptive degradations such as Gaussian and shot noise. For corruptions that minimally affect apparent motion (\textit{e.g.,} contrast shifts), semantic fidelity becomes relatively more important. Notably, we observe that substantial increases in warp error consistently lead to pronounced navigation degradation—even when semantic restoration appears adequate (\textit{e.g.,} Dec-DPM under JPEG compression and Dec-CM under fog).

Inference latency remains a pivotal consideration for online VLN deployment. As reported in Table~\ref{tab:time}, FlowDec achieves single-step decorruption in merely 169 ms—delivering 3$\times$ to 8$\times$ speedup over Dec-DPM and Dec-CM. With episodes typically comprising $\sim$70 steps, the cumulative overhead of diffusion baselines renders them infeasible for real-time operation. FlowDec, uniquely, reconciles stringent robustness requirements with low-latency constraints, establishing it as a practical solution for embodied navigation in adverse visual conditions. 

We further assess FlowDec's impact under pristine conditions. Results in Table~\ref{tab:clean} confirm that its integration incurs no statistically significant performance degradation on clean inputs. The minor degradation in clean unseen scenes is attributed to the domain gap between synthetic training corruptions and clean images~\cite{tsai2024gda,guo2025everything}. However, the trade-off is small compared to the statistically significant gains under corruption, which is the primary target scenario. 
\begin{figure}[t]
  \centering
   \includegraphics[width=1\linewidth]{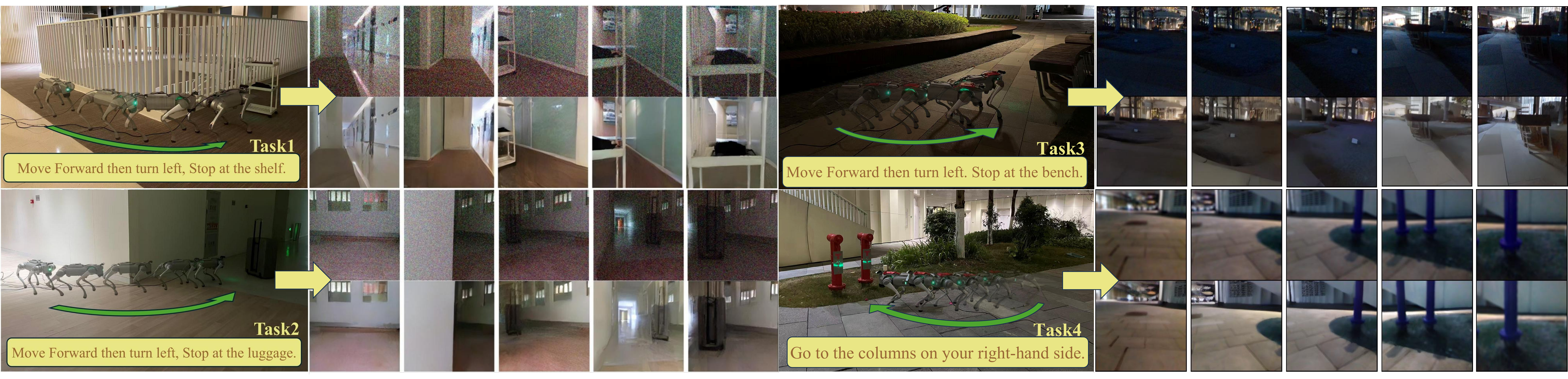} %还有另一个版本见fig51.pdf
   \caption{Illustration of experiments in real-world environment. The two rows of images on the right side represent the original image and the decorrupted image respectively.}
   \label{fig5}
\end{figure}

\subsection{Deployment on Legged Robots.} %\textcolor{blue}{To evaluate the effectiveness of our model in real world scenarios, we designed four representative tasks containing both indoor and outdoor scenarios. The indoor tasks are added Gaussian noise while the outdoor tasks contain worse luminance condition and motion blur. As reported in Table~\ref{tab:real}, FlowDec consistently outperforms the original baseline model across all four tasks. Notably, the performance gains are more pronounced in the more challenging scenarios with stronger luminance variation (\textit{e.g.,} Task 2 and 3). These results validate FlowDec’s generalization to unseen real-world conditions and can adapt well to common real-world interference.}

To validate the effectiveness of our approach in real-world settings, we deploy our system on a legged robot and design four representative navigation tasks spanning both indoor and outdoor environments. The indoor tasks are conducted under synthetic Gaussian noise to simulate sensor degradation, while the outdoor tasks involve more severe luminance variations and motion blur caused by dynamic lighting and ego-motion.
As reported in Table~\ref{tab:real}, FlowDec consistently outperforms the original baseline across all four tasks. Notably, the performance improvements are more substantial in the more challenging scenarios with pronounced luminance fluctuations (\textit{e.g.,} Task 2 and Task 3), where perceptual degradation significantly impacts navigation reliability. These results demonstrate that FlowDec generalizes effectively to previously unseen real-world conditions and robustly adapts to common environmental and sensor-induced disturbances.

\subsection{Ablation Studies}
\label{sec:4.5}

\begin{figure}[t]
  \centering
   \includegraphics[width=1\linewidth]{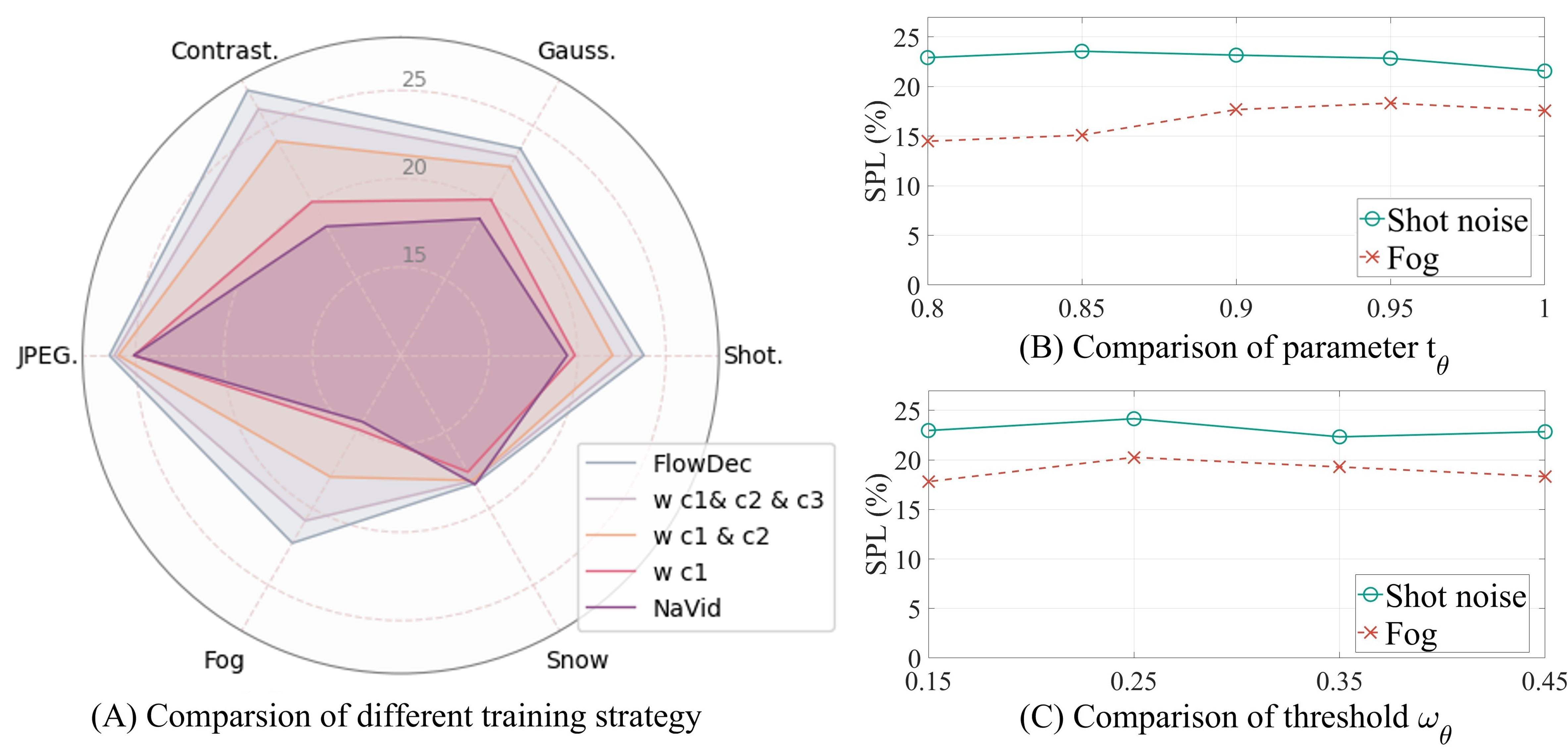}
   \caption{Summary of ablation study results. All results are measured by SPL.}
   \label{fig6}
\end{figure}

\textbf{Training strategy.} The primary goal of incorporating temporal conditions ($\mathbf{c}_2$, $\mathbf{c}_3$) during training is to align the model’s flow path with the primary condition ($\mathbf{c}_1$). Since all conditioning objectives are mutually consistent, auxiliary conditions enhance optimization of the main flow, improving $\mathbf{c}_1$ robustness even without temporal input at inference. To validate this, we evaluate models trained under different strategies on R2R-CE, using only $\mathbf{c}_1$ (without filtering) during inference. SPL results (Figure~\ref{fig6}(A)) show universal gains after introducing the Temporal Condition Training Strategy. These improvements stem from enhanced spatial consistency: aligned flow paths produce coherent object representations across frames, even in single-frame inference. The term `w $\mathbf{c}_1 \& \mathbf{c}_2 \& \mathbf{c}_3$' refers to the FlowDec training strategy without action-centroid guided latent filtering.

\textbf{Various starting points.}  To accelerate inference, we modify the flow matching starting point via $t_\theta$, initializing from a linear interpolation between random noise and the corrupted latent. While this reduces denoising capacity, it enhances consistency with the input image. Figure~\ref{fig6}(B) confirms this trade-off: on shot noise and fog, SPL is lower when starting from pure noise ($t_\theta = 1$) than from an intermediate state. A small proportion of corrupted input ($t_\theta = 0.95$) significantly improves performance,  whereas a larger proportion diminishes the ability of the model to decorruption, leading to a gradual decline in effectiveness.

\textbf{Filtering threshold.} The filtering threshold $w_\theta$ controls the contribution of the auxiliary condition $\mathbf{c}_3$ based on Mahalanobis distance to the action centroid, enabling adaptive use of historical context for spatial consistency. Across corruptions, this distance ranges from 0.15 to 0.45. Figure~\ref{fig6}(C) evaluates shot noise and fog measured by SPL: moderate integration ($w_\theta \geq 0.25$) optimally balances temporal guidance and primary condition fidelity, maximizing SPL. While excessive reliance on $\mathbf{c}_3$ ($w_\theta < 0.25$) degrades performance as well as increases unnecessary inference times.

\section{Conclusion}
VLN plays a fundamental role in embodied AI and continues to attract growing attention. In this work, we address the vulnerability of existing LM-based VLN models to visual corruptions by proposing the FlowDec, a novel module designed to enhance robustness. Through its hybrid temporal conditioning strategy and action-centroid guided filtering, FlowDec effectively mitigates the impact of unseen corruptions while preserving temporal consistency across generated images. This results in substantial performance improvements in both simulated and real-world environments. Moreover, the rapid inference speed of FlowDec supports real-time navigation, making it highly practical for online deployment. Looking ahead, we plan to extend this framework by developing denoising models robust to dynamically varying noise patterns, further advancing the resilience and applicability of VLN systems in more complex real-world scenarios.

\section*{Acknowledgments}
This work was supported by National Natural Science Foundation of China (NFSC) under the Grant Number 62573370 and Key Area Project of Education Department of Guangdong Province (No. 2025ZDZX3051).

\bibliographystyle{splncs04}
\bibliography{main}
\clearpage
\appendix

\section{Pseudo-code for FlowDec}
\label{sec:rationale}
The \textbf{pseudocode} for the FlowDec is shown in two phases.

\begin{itemize}
    \item \textbf{Training} (Algorithm~\ref{alg:train}): In the \textbf{first epoch}, \textbf{action centroids} are constructed from ground-truth image pairs. Subsequent epochs train the model using \textbf{augmented images} and a \textbf{hybrid temporal conditioning strategy} with dynamic condition weighting.
    
    \item \textbf{Inference} (Algorithm~\ref{alg:infer}): The model \textbf{iteratively decorrupts} input frames, guided by \textbf{VLN-predicted actions}. \textbf{Action-centroid guided filtering} selectively invokes temporal conditioning only when deviations exceed threshold $w_\theta$, ensuring \textbf{consistency} and \textbf{efficiency}.
\end{itemize}

\begin{algorithm}
\caption{Training Phase of FlowDec}
\label{alg:train}
\begin{algorithmic}[1]
\REQUIRE Current and previous images, ground-truth actions $a_p$
\STATE Encode current and prior ground-truth images via VAE encoder $\mathcal{E}$~\cite{Kingma_Welling_2013} to obtain $\mathbf{z}_{\text{c\_gt},1}$, $\mathbf{z}_{\text{p\_gt},1}$
\STATE Construct \textbf{action centroids} using Eq.~\eqref{eq11}
\STATE Initialize hybrid condition weights
\FOR{each training epoch}
    \STATE Apply augmentation~\cite{oh2024efficient} to generate $\mathbf{z}_{\text{c\_aug},1}$, $\mathbf{z}_{\text{p\_aug},1}$
    \STATE Sample condition $\mathbf{c}_i \sim$ current distribution
    \STATE Compute CFM loss via Eq.~\eqref{eq9} and update model
    \STATE Update condition weights: increase $\mathbf{c}_3$, decrease $\mathbf{c}_2$
\ENDFOR
\end{algorithmic}
\end{algorithm}

\begin{algorithm}
\caption{Inference Phase of FlowDec}
\label{alg:infer}
\begin{algorithmic}[1]
\REQUIRE Corrupted image stream, predicted actions $a_p$ from VLN model
\WHILE{action $\neq$ ``STOP''}
    \STATE Compute starting latent $\mathbf{z}'_0$ via Eq.~\eqref{eq14}
    \STATE Generate $\mathbf{z}_{\text{c\_decor}}$ using $\mathbf{c}_1$
    \IF{not first frame}
        \STATE Compute Mahalanobis distance $w$ to action centroid (Eq.~\eqref{eq12})
        \IF{$w > w_\theta$}
            \STATE Generate $\tilde{\mathbf{z}}_{\text{c\_decor}}$ using $\mathbf{c}_3$, compute $\tilde{w}$
            \STATE Integrate latent via Eq.~\eqref{eq13}.
        \ENDIF
    \ENDIF
    \STATE Decode final latent via $\mathcal{D}$ and forward to VLN model
\ENDWHILE
\end{algorithmic}
\end{algorithm}

\section{Benchmarks, Backbone and Training Details}
\subsection{Benchmark}
\begin{figure}[t]
  \centering
   \includegraphics[width=1\linewidth]{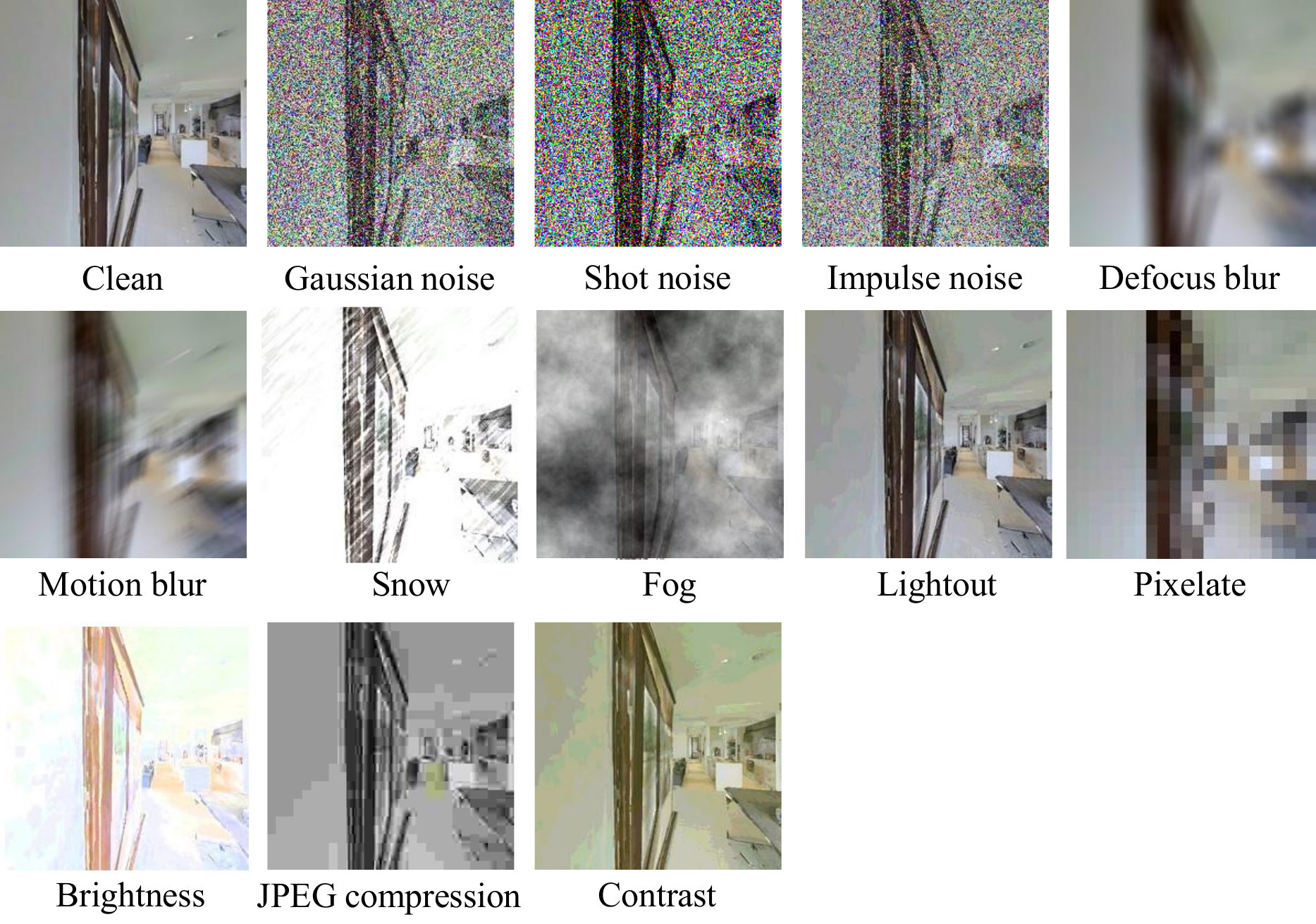}
   \caption{Visualization of 12 corruption types on the sampled image in R2R-CE benchmark.}
   \label{fig7}
\end{figure}

All experiments are conducted on two prominent VLN-CE datasets: R2R-CE \cite{krantz2020beyond} and RxR-CE \cite{ku2020room}. To assess robust navigation under visual degradations, we introduce corruptions to the visual modality. Following \cite{wang2022continual, oh2024efficient}, we apply 12 corruption types for comprehensive evaluation: Gaussian noise, shot noise, impulse noise, defocus blur, motion blur, zoom blur, snow, fog, light-out, pixelate, JPEG compression, contrast, and brightness. The corrupted variants of these benchmarks are referred to as R2R-CE-Corrupt and RxR-CE-Corrupt, respectively. Visualizations of sample corrupted video frames are provided in Figure~\ref{fig7}.

\subsection{Backbone}
NaVid is used as the backbone VLN network in this paper. NaVid, built on the LLaMA-VID framework, integrates Vicuna-7B as the core LLM for instruction-following and action reasoning, EVA-CLIP as the frozen vision encoder, BERT as the text encoder, and a Q-Former-based query generator that produces instruction-aware queries via cross-attention with the instruction embeddings. Two cross-modality projectors align visual tokens into the LLM’s language space. NaVid is trained on 1.2M data including action planing, image QA, video QA, \textit{etc.} For input, NaVid processes a monocular RGB video stream: historical frames are each encoded into 1 instruction-queried token and 4 instruction-agnostic tokens, while the current frame uses 64 instruction-agnostic tokens for fine-grained geometry; all are concatenated with instruction tokens and structured using special tokens to distinguish history, current observation, and navigation prompt. The output is a linguistic action string specifying one of four types — FORWARD, TURN-LEFT, TURN-RIGHT, STOP — with quantitative arguments, which a regex parser extracts for direct low-level robot execution in continuous environments.

\subsection{Training Details}
The training dataset for FlowDec is derived from R2R-CE and RxR-CE. The R2R-CE dataset has 10,819 training and 1,839 validation episodes, while RxR-CE has 19,561 training and 3,669 validation episodes. We collect RGB image pairs along agent trajectories from the training episodes in the Habitat simulator~\cite{savva2019habitat}, stored in the format:
\[
[\text{current action } a_p,\ \text{current image},\ \text{previous image}]
\]
Five \textbf{atomic actions} are included:
\begin{itemize}
    \item \textbf{Move Forward 25 cm} (1,628,366 samples)
    \item \textbf{Turn Right 15°} (110,882 samples)
    \item \textbf{Turn Left 15°} (104,072 samples)
    \item \textbf{Turn Right 30°} (307,582 samples)
    \item \textbf{Turn Left 30°} (289,912 samples)
\end{itemize}
These correspond to executable actions in R2R-CE and RxR-CE.

For the hybrid temporal conditioning strategy, the weight of $\mathbf{c}_1$ is fixed at 0.3. The remaining 0.7 is dynamically allocated between $\mathbf{c}_2$ and $\mathbf{c}_3$:
\[
w(\mathbf{c}_2) = 0.7 \times (1 - 0.8 \times \frac{\text{cur\_epoch}}{\text{total\_epochs}}), 
\]
\[
w(\mathbf{c}_3) = 0.7 \times 0.8 \times \frac{\text{cur\_epoch}}{\text{total\_epochs}}.
\]
As described in Section~\ref{sec:3.2}, training initially prioritizes $\mathbf{c}_2$ to leverage ground-truth priors for stable corruption modeling. As the model converges, $\mathbf{c}_3$ dominates to bridge the synthetic-to-real domain gap.
All other hyperparameters follow~\cite{dao2023flow}, except those ablated in Section~\ref{sec:4.5}. The model is trained for 20 epochs and the initial learning rate is set to $5e-5$.
\newpage
\section{More Experiment Results}

\subsection{Visualization of Action-Centroid Guided Filtering.}
\begin{figure}[t]
  \centering
   \includegraphics[width=0.9\linewidth]{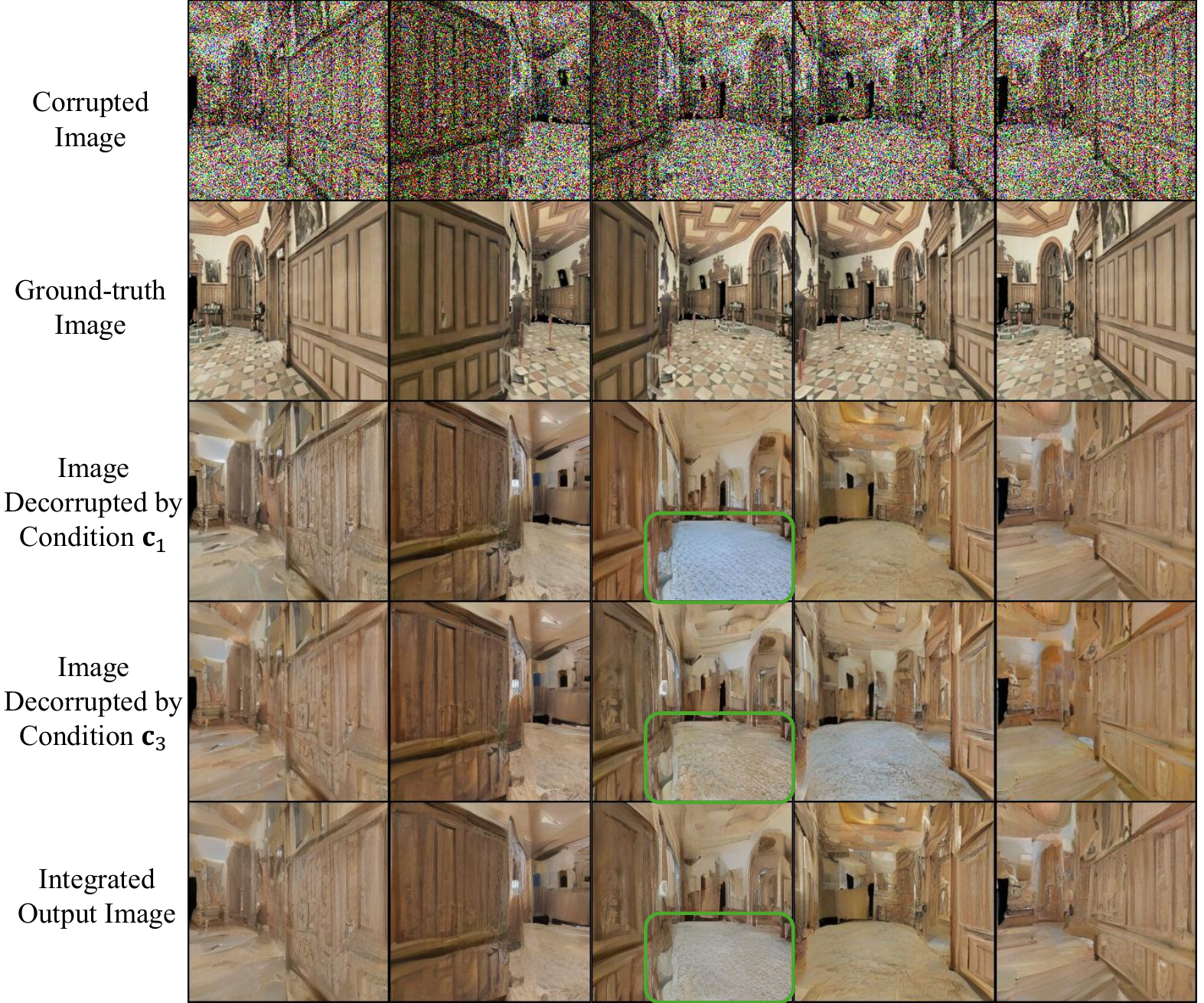}
   \caption{Visualization of action-centroid guided filtering.}
   \label{fig8}
\end{figure}

Figure~\ref{fig8} presents a navigation sequence denoised under fog corruption. For each frame, we show outputs from $\mathbf{c}_1$, $\mathbf{c}_3$, and the integrated result. In most frames, the integrated image is identical to the $\mathbf{c}_1$ output, confirming that FlowDec relies primarily on the stable single-frame pathway. Only when severe inter-frame inconsistency occurs (\textit{e.g.,} drastic floor color shift highlighted in green) does the action-centroid distance exceed $w_\theta$, triggering fusion with $\mathbf{c}_3$. This selective mechanism effectively corrects temporal artifacts while minimizing unnecessary computation.

\begin{figure}[htp]
    \centering
    \includegraphics[width=0.9\linewidth]{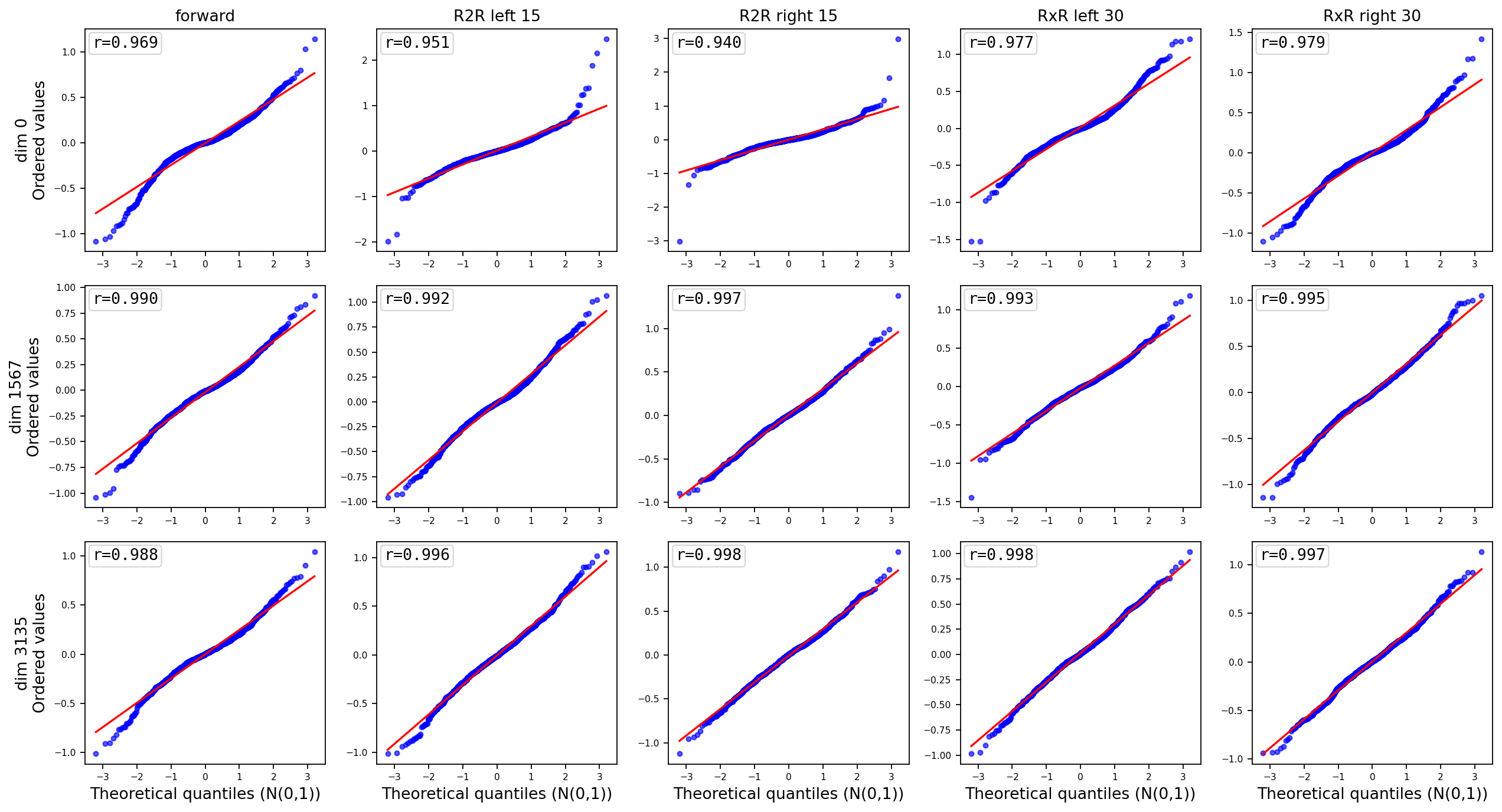}
    \caption{Q-Q plots of differential latents for each action type.}
    \label{Q-Q}
\end{figure}

To validate the Gaussian assumption, we sampled 1000 differential latents per action, reduced them to three dimensions via PCA, and plotted Q-Q plots in Figure~\ref{Q-Q}. The Q-Q correlation $r > 0.95$ across all dimensions confirms that the Gaussian fit is practically adequate.

\subsection{Performance on other benchmark.}
In this section, we provide additional results using another backbone, Uni-NaVid~\cite{zhang2024uni}, to further evaluate our model's effectiveness. Uni-NaVid is a video-based model that unifies multiple embodied navigation tasks within a single framework, enabling general-purpose, fast, and long-horizon navigation in unseen real-world environments. The results in Tables~\ref{tab:uni-corruption} and~\ref{tab:uni-compare} show the same trend as discussed in Sections~\ref{sec:4.2} and~\ref{sec:4.3}. This occurs because most transformer-based VLN-CE systems primarily focus on contributions such as obstacle avoidance or multi-task unification, which are orthogonal to the problem addressed in our study. This demonstrates that FlowDec is a model-agnostic, plug-and-play visual module that does not depend on specific backbone designs. 

\begin{table*}[t]
\centering
\caption{Performance of Uni-NaVid under different corruption types on R2R-CE and RxR-CE datasets. All metrics are reported in \%.}
\label{tab:uni-corruption}
\resizebox{1\linewidth}{!}{
\begin{tabular}{cc*{14}{c}}
\toprule
Dataset & Metric & Clean & Gauss. & Shot & Impul. & Defoc. & Motion. & Snow & Fog & Lightout & Pixel. & JPEG. & Cont. & Bright. \\
\midrule
\multirow{2}{*}{R2R-CE} 
 & SR  &51.80&31.87&29.36&26.86&31.20&34.64 &30.71&31.05&37.03&37.75&34.66&38.28&36.92  \\
 & SPL & 47.70&30.58&28.27&25.30&29.81&32.98&29.27&29.56&35.46&36.18&32.46&36.26&34.85 \\
\midrule
\multirow{2}{*}{RxR-CE} 
 & SR  & 56.1&34.78&32.92&30.36&31.64&36.15&30.77&34.56&43.64&44.93&41.62&41.62&44.09 \\
 & SPL & 44.5&29.89&27.88&25.81&26.47&31.67&25.43&29.54&36.27&38.64&34.65&34.62&38.18 \\
\bottomrule
\end{tabular}}\vspace{-0.5em}
\end{table*}

\begin{table*}[t]
\centering
\caption{Performance of FlowDec under corruption on R2R-CE and RxR-CE. All metrics are reported in \%.}
\label{tab:uni-compare}
\sisetup{
  round-mode = places,
  round-precision = 2,
  zero-decimal-to-integer = false,
  group-digits = false,
}
\newcommand{\best}[1]{%
  \ifdim #1pt=\maxofcol pt \textbf{#1}\else #1\fi
}
\resizebox{\linewidth}{!}{%
\begin{tabular}{
  l c
  *{7}{S[table-format=2.2]}
  >{}S[table-format=2.2]
  c@{\hspace{1em}}
  *{7}{S[table-format=2.2]}
  >{}S[table-format=2.2]
}
\toprule
\multirow{2}{*}{Method} & \multirow{2}{*}{Metric}
 & \multicolumn{7}{c}{R2R-CE} && \multicolumn{7}{c}{RxR-CE} \\
\cmidrule(lr){3-9} \cmidrule(lr){10-17}
 & & {Gauss.} & {Shot}  & {Snow} & {Fog} & {JPEG.} & {Cont.}& {Avg.}
 & & {Gauss.} & {Shot} & {Snow} & {Fog} & {JPEG.} & {Cont.}& {Avg.} \\ \midrule
\multirow{2}{*}{Uni-NaVid}
 & SR &31.87&29.36&26.86&31.20&34.64&30.34&30.71&&34.78&32.92&30.36&31.64&36.15&30.77&32.77 \\
 & SPL&30.58&28.27&25.30&29.81&32.98&28.67 &29.27&&29.89&27.88&25.81&26.47&31.67&25.43&27.86  \\
\addlinespace
\multirow{2}{*}{FlowDec} 
& SR  &32.72&30.20&26.89&33.66&35.90&35.56&32.49&&35.36&34.79&30.46&35.08&36.92&40.36&35.50 \\
& SPL &30.88&28.95&25.16&29.92&33.20&34.00&30.35&&30.17&29.54&25.73&30.52&30.85&35.82&30.44 \\
\bottomrule
\end{tabular}%
}\vspace{-1em}
\end{table*}

\subsection{Solving Steps of FlowDec.}
\begin{figure}[hbt]
\vspace{-0.5em}
  \centering
   \includegraphics[width=0.9\linewidth]{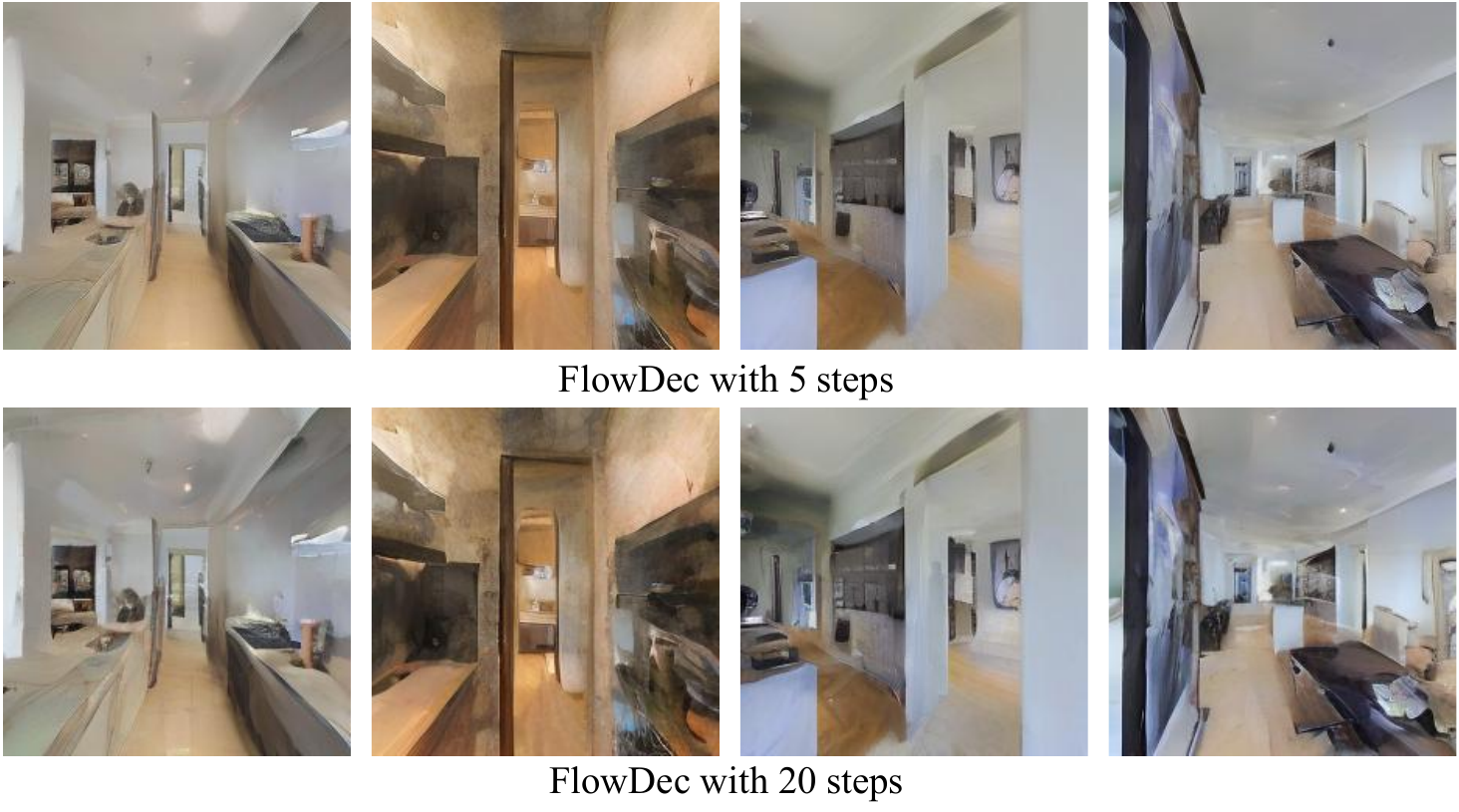}
   \caption{Visualization of images denoised with different steps, using fog as the corruption type.}
   \label{fig9}
\end{figure}

FlowDec uses a 20-step Euler solver by default. To evaluate the impact of fewer steps, we compare 5-step, 10-step, and 20-step inference under identical $t_\theta$ and $w_\theta$ on R2R-CE (Table~\ref{tab:supl}). Although the 20-step version achieves the highest average SPL, the gains are marginal. Surprisingly, 5-step inference occasionally outperforms 20 steps (\textit{e.g.,} Gaussian noise and fog), suggesting that excessive refinement can introduce minor domain drift irrelevant to navigation. Visual comparison (Figure~\ref{fig9}) confirms negligible perceptual differences; 20-step results offer only subtle details improvement that provide limited benefit to VLN-CE. Thus, reducing inference steps to 5–10 represents an effective speed–accuracy trade-off.

\begin{table*}[ht]
\centering
\small
\caption{Performance of different methods under different solution steps. All metrics are reported in \%. The best results for each domain are highlighted in \textbf{bold}.}
\label{tab:supl}
\sisetup{
  round-mode = places,
  round-precision = 2,
  zero-decimal-to-integer = false,
  group-digits = false,
}
\newcommand{\best}[1]{%
  \ifdim #1pt=\maxofcol pt \textbf{#1}\else #1\fi
}
\resizebox{0.8\linewidth}{!}{%
\begin{tabular}{
  l c
  *{8}{S[table-format=2.2]}
  >{}S[table-format=2.2]
}
\toprule
Method & Metric & {Gauss.} & {Shot} & {Snow} & {Fog} & {Jpeg.} & {Cont.}& {Avg.} \\ \midrule
\addlinespace
\multirow{2}{*}{FlowDec-5}
 & SR & \textbf{26.86}&	25.67&21.21&	\textbf{24.20}&	28.98&	28.27&	25.88  \\
& SPL& \textbf{24.18}&	22.88&	\textbf{18.47}&	\textbf{21.32}&	26.17&	24.27&	22.88  \\
\addlinespace
\multirow{2}{*}{FlowDec-10}
& SR & 26.43&	25.56&	19.25&	22.24&	27.70&	26.26&	24.57 \\
& SPL & 23.30&	22.62&	16.69&	19.19&	23.63&	23.02&	21.41 \\
\addlinespace
\multirow{2}{*}{FlowDec-20} & SR & 26.54 & \textbf{26.70} &\textbf{ 21.40} & 23.27 & \textbf{29.85} &\textbf{ 30.90} &  \textbf{27.04} \\
& SPL & 23.53 & \textbf{23.76} & 18.38 & 20.26 & \textbf{26.48} & \textbf{27.32 }  & \textbf{23.62} \\
\bottomrule
\end{tabular}%
}
\vspace{-0.5em}
\end{table*}

\end{document}